\documentclass{article}

\PassOptionsToPackage{numbers, compress}{natbib}

\usepackage[eandd, preprint]{neurips_2026}

\usepackage[utf8]{inputenc} %
\usepackage[T1]{fontenc}    %
\usepackage{hyperref}       %
\usepackage{url}            %
\usepackage{booktabs}       %
\usepackage{amsfonts}       %
\usepackage{nicefrac}       %
\usepackage{microtype}      %
\usepackage[table]{xcolor}  %
\usepackage{amsmath}
\usepackage{tabularx}
\usepackage{booktabs}  %
\usepackage{multirow}  %
\usepackage{makecell}
\usepackage{graphicx}
\usepackage{enumitem}
\usepackage{graphicx}    %
\usepackage{subcaption}  %
\usepackage{tcolorbox}

\title{What-If World: A Causal Benchmark for General World Models in Embodied Scenarios}

\author{
{\large\bfseries Kunlin Cai\textsuperscript{1} \quad Rui Song\textsuperscript{1} \quad Jinghuai Zhang\textsuperscript{1} \quad Kaiyuan Zhang\textsuperscript{1}}\\[3pt]
{\large\bfseries Pranav Bodapati\textsuperscript{1} \quad Alicia Yu\textsuperscript{1} \quad Fnu Suya\textsuperscript{2}}\\[3pt]
{\large\bfseries Mohammad Rostami\textsuperscript{3}\thanks{Work is independent from position at Amazon.} \quad Jiaqi Ma\textsuperscript{1} \quad Yuan Tian\textsuperscript{1}}\\[6pt]
\textsuperscript{1}UCLA \quad \textsuperscript{2}University of Tennessee \quad \textsuperscript{3}Amazon \\[4pt]
\texttt{\{kunlin96, rruisong, jinghuai1998, kaiyuanzhang, pranavbodapati1,}\\
\texttt{aliciayu, jiaqima, yuant\}@g.ucla.edu} \\[2pt]
\texttt{fsuya@utk.edu} \quad \texttt{rostamii@amazon.com}
}

\begin{document}

\maketitle

\begin{abstract}

Video generation models are increasingly used as world simulators for tasks like driving and robotic manipulation. What matters in these settings is not whether a single video looks right, but whether the model's output changes when its input changes. We test this by giving a model two prompts describing the same scene with one physical detail varied, and checking whether the two videos diverge the way physics predicts. The wording difference between the prompts is small by design, since only one variable is changed, but the correct physical difference is not. A model that misses this can still produce two videos that each look plausible individually, and existing benchmarks score videos one at a time and cannot detect this failure. We introduce What-If World, 319 such prompt pairs built on real frames from nuScenes and DROID, organized by a taxonomy of six physical variables shared across driving and manipulation. Each pair is scored with APEO, a four-part rubric checking whether each video follows its prompt (Adherence), is physically consistent (Physics), preserves the shared scene (Environment), and ends in the correct difference (Outcome). Across nine state-of-the-art models, no system exceeds 52\% on the paired score, and open-source models cluster near 28\%. Every model tested fails on a large fraction of causal interventions, indicating substantial room before these models can reliably support action-conditioned simulation or model-based planning. Where models do score well, performance appears to track the visual prominence of the intervention rather than the tractability of its underlying physics. Some visually subtle interventions score as low as 14.2\%, while visually pronounced ones reach 40.4\%.

\end{abstract}

\section{Introduction}
\label{sec:intro}

Video generation models are increasingly framed as general-purpose world simulators for embodied applications such as autonomous driving and robotic manipulation~\cite{deepmind2025genie3, openai2024worldsimulators}. In these settings, visual realism is typically not enough, and neither is isolated physical plausibility~\cite{bardes2024revisiting, hu2023gaia}. A model can render a car braking hard with flawless kinematics yet produce nearly identical footage when the prompt is changed to a gentler brake. It has captured the visual template of ``a car braking'' but remains insensitive to the input that should shape the outcome. For action-conditioned simulators, model-based planners, and policy training pipelines, this is the failure that matters most, since rollouts that fail to track the controlled variable cannot be used to compare alternative actions or conditions.

This failure mode is inherently comparative: a single video rarely reveals whether the model would have produced the same footage under a different input. Yet existing benchmarks evaluate each video in isolation. Video-quality benchmarks such as VBench~\citep{DBLP:conf/cvpr/HuangHYZS0Z0JCW24} and EvalCrafter~\citep{DBLP:conf/cvpr/LiuC0WZCLZCS24}, together with physics-focused extensions like VideoPhy~\citep{bansal2024videophy} and PhyGenBench~\citep{DBLP:conf/icml/MengLTLSZ0L025}, score each generation against a single prompt, so two near-identical videos for "brake gently" and "brake hard" would both pass. Causal benchmarks for visual understanding, including CLEVRER~\citep{DBLP:conf/iclr/YiGLK0TT20} and NExT-QA~\citep{DBLP:conf/cvpr/XiaoSYC21}, instead test whether a model perceives causal structure in a video shown to it—not whether the videos it generates track changes in its input. Neither line of work probes the property at stake: that a controlled change in input should yield a correspondingly controlled change in output.

Probing this property in video generation models requires
formalization across three pieces, namely benchmark dimensions,
construction pipeline, and evaluation protocol. First, no prior
work defines what counts as a valid physical intervention
variable in embodied video generation, so the test space must
be established before any selection. Second, controlled
comparison is fragile in a generative setting, where the model
is text-driven yet must produce a physical effect, and \emph{ceteris
paribus} must be approximated through input design rather than
enforced through generation. Third, the evaluation faces two
intertwined difficulties. Causal sensitivity is not a single
property, since a pair can fail in several ways. The intervention
may not have executed, the videos may execute but not diverge,
or the divergence may reflect scene drift rather than the
intervention. Without separating these modes, the evaluation
cannot tell what went wrong. Moreover, these judgments require
semantic reasoning that feature-distance metrics cannot capture,
while the benchmark's scale rules out human annotation.

To fill these gaps, we build \textbf{What-If World}, a benchmark for contrastive intervention in autonomous driving and robotic arm manipulation. For test dimensions, two domain experts apply the three criteria above to real videos and merge survivors by causal mechanism (e.g., braking and acceleration intensity collapse into a single Force/Degree primitive), yielding six primitives across two domains: environment (surface friction, material/medium, obstacle configuration) and agent action (spatial alignment, force/degree, temporal sequencing). For test constructions, each instance is built in three steps.
(1) We filter clips from nuScenes~\citep{caesar2020nuscenes}
or DROID~\citep{khazatsky2024droid} for the \emph{physical
conditions} the target intervention needs. (2) We extract
the frame just before the action begins and pair it with two scene-describing prompt
components (camera perspective and initial scene state),
together anchoring the shared initial state. (3) We author
a prompt pair that differs in exactly one semantic variable, so that any divergence between the two videos
traces back to the intervention. 
For evaluations, we introduce the \textbf{APEO} framework, scoring each pair on Adherence (intervention executed), Physics (dynamics valid), Environment (scene held still), and Outcome (result diverged as predicted); the first three are scored in both single-video and paired modes, while Outcome is paired-only. The rubric runs at benchmark scale via a VLM judge validated against human annotation.

Applying this protocol to nine state-of-the-art video generation world models (four open-source, five closed-source) with Gemini 3.1-Pro as the VLM judge, we report three findings. First, causal sensitivity in video generation. No model exceeds 52\% on our paired metric, and the four open-source models cluster near 28\%. Second, most models show a wide gap between per-video and paired evaluation scores, generating pairs of individually convincing videos that fail to differ in the way the input predicts. We call this gap the contrastive bottleneck. Third, where models do score well, the pattern is more consistent with the visual prominence of the intervention's consequences than with generalizable physical sensitivity—Force/Degree (producing dramatic motion differences) reaches 40.4\% while Surface Friction (producing subtle trajectory differences) drops to 14.2\%.

\noindent In summary, our contributions are as follows.

\begin{enumerate}[leftmargin=*, noitemsep, topsep=0pt]
    \item \textbf{The What-If World benchmark}, the first contrastive-intervention benchmark for video world models in embodied domains, contributing a six-primitive taxonomy of physical intervention variables that unifies autonomous driving and robotic manipulation, instantiated as 319 paired triplets anchored on real frames from nuScenes and DROID.
    \item \textbf{The APEO evaluation framework}, a paired protocol scoring Adherence, Physics, Environment, and Outcome in both single-video and paired modes, with a VLM-based implementation validated against human annotation.
    \item \textbf{A diagnostic study of nine state-of-the-art video generation models}, which surfaces the failure modes characterized above and provides the first quantitative evidence that they are invisible to existing per-video evaluation.
\end{enumerate}

\section{Background and Related Work}

\textbf{Video generation benchmarks cannot test causal sensitivity.}
Existing benchmarks for video generation measure a wide range of properties. One line of work focuses on visual quality~\citep{DBLP:conf/cvpr/HuangHYZS0Z0JCW24, DBLP:conf/cvpr/SunHL0XLL25, zheng2025vbench2, huang2025vbench++}. Others assess 3D consistency~\cite{duan2025worldscore, zhen2025tesseract, wu2025video}, the physical plausibility of individual generated clips~\citep{bansal2024videophy, DBLP:conf/icml/MengLTLSZ0L025, bansal2025videophy, li2025worldmodelbench}, or spatial--temporal coherence~\cite{yuan2024chronomagic, chen2025learning}. More recently, a separate line of work evaluates video world models on embodied downstream tasks such as autonomous driving and robotics~\cite{zhou2026drivinggen, qin2024worldsimbench, shang2026worldarena}. None of these benchmarks, however, probes \emph{causal sensitivity}: whether a model's output changes when the prompt is intervened on. This gap matters because visual realism is a poor proxy for physical correctness~\cite{motamed2026generative, DBLP:conf/icml/KangYL00W0F25}: current generators routinely produce convincing footage that violates the underlying physics, undermining the physical fidelity that embodied applications require~\cite{gupta2024essential, wang2025omnidrive}. Consequently, a model that returns the same video for ``light braking'' and ``hard braking'' would pass every existing benchmark undetected.

\noindent \textbf{Causal benchmarks cannot test video world models.}
VQA benchmarks have been used to evaluate the ability of models to understand causal or physical phenomena~\cite{niu2021counterfactual, DBLP:conf/iclr/YiGLK0TT20, DBLP:conf/cvpr/XiaoSYC21, grunde2021agqa}. The recent evaluation trend has shifted toward large language models~\cite{wang2024causalbench, chi2024unveiling, maasch2025compositional} and vision-language models~\citep{wu2024star, DBLP:journals/corr/abs-2506-09943, parmar2024causalchaos, komanduri2025causalvlbench, li2024multimodal}, which present real visual inputs and require models to answer causal questions with discrete labels. Moreover, image editing benchmarks~\citep{DBLP:conf/cvpr/BrooksHE23, DBLP:conf/nips/ZhangMCSS23} take a real reference photograph and measure whether a spatial change has been applied correctly---no temporal dynamics are required, and the evaluation simply compares the output to the original. However, with the recent emergence of generative world models~\cite{hafner2023mastering, alonso2024diffusion, wang2024drivedreamer, gao2024vista}, both paradigms prove insufficient, since neither requires a model to \emph{generate} physically correct causal dynamics from scratch. What-If World addresses this gap: starting from the same real initial frame, the model must generate two videos that differ physically in precisely the manner predicted by the intervention on the causal variable.

\section{The What-If World Benchmark}

What-If World aims to benchmark a specific capability: whether a video generation model can simulate the differential physical consequences of a controlled intervention, rather than merely render visually plausible footage. Formally, a model $\mathcal{M}$ takes an initial frame $x_0$ and a text prompt $p$ to generate a video $\mathcal{V} = \mathcal{M}(x_0, p)$. We define the capability under test as \textit{causal intervention} as follows: given a shared $x_0$ and a pair of prompts $(p^+, p^-)$ that differ only in the description of one physical variable $v$, the model $\mathcal{M}$ should produce two videos $(\mathcal{V}^+, \mathcal{V}^-)$ that diverge in the direction physics predicts under the intervention. Building this benchmark requires answering three questions:

\begin{itemize}[leftmargin=*, noitemsep, topsep=0pt]
    \item \textbf{What to test} (Section~\ref{sec:taxonomy}): which physical variables to put under intervention. We construct a six-primitive taxonomy that spans agent actions and environment conditions, retaining only variables that are physically fundamental, embodiment-shared, and operationally isolable.
    
    \item \textbf{How to test} (Section~\ref{sec:construction}): how to construct each test pair so the  divergence can be attributable to the intervened variable. We anchor every test on a real frame $x_0$ at the causal branching point and write a contrastive prompt pair that differs only in the description of the control.
    
    \item \textbf{How to evaluate} (Section~\ref{sec:evaluation}): how to score the pair jointly. Our APEO framework checks four dimensions (Adherence, Physics, Environment, and Outcome) in both single-video and paired modes, so per-video plausibility is separated from causal correctness.
\end{itemize}

\begin{figure}[tbp]
    \centering
    \includegraphics[width=1\textwidth]{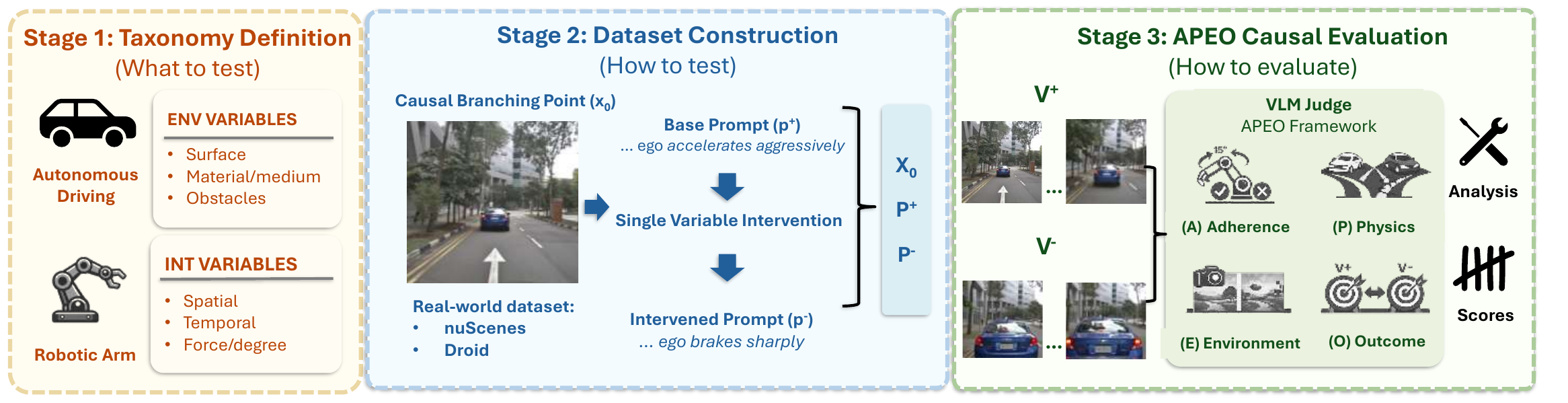}
    \caption{What-If World benchmark overview. Three stages address the challenges of evaluating the causal sensitivity of video world models: (1) what to test, (2) how to test, and (3) how to evaluate.}
    \label{fig:overview}
\end{figure}

\subsection{The What-If World Causal Taxonomy}
\label{sec:taxonomy}

To address the \textbf{What to test} challenge, we need to identify the variables that govern physical outcomes. A taxonomy of such variables is not just a descriptive convenience but a prerequisite of our evaluation protocol for two reasons: (1) The contrastive pair $(p_+, p_-)$ is only well-defined once we name the variable being changed; without an explicit set of variables to draw from, ``changing one physical variable'' has no operational meaning. (2) Aggregate scores become actionable diagnostics only when each failure can be traced to a specific variable. A taxonomy lets us report which class of physical phenomena a model fails on, rather than a single opaque ``physics'' score.

As discussed in the background, existing benchmarks do not supply such a space. We therefore construct a taxonomy of physical primitives, grounded in real-world data~\citep{braun2006using}. Two domain experts (one in AD, one in manipulation) randomly selected and reviewed 1,000 real videos from nuScenes~\citep{caesar2020nuscenes} and DROID~\citep{khazatsky2024droid} to identify variables that change physical outcomes. Candidate variables were retained only if they were: (i) \textit{physically fundamental}, driven by a real physical mechanism rather than a stylistic or perceptual change~\citep{pearl2009causality, DBLP:conf/icml/KangYL00W0F25}; (ii) \textit{embodiment-shared}, applicable to both driving and robotics so that performance reflects general physical sensitivity rather than domain-specific shortcuts~\citep{o2024open,DBLP:journals/corr/abs-2506-09943}; and (iii) \textit{operationally isolable}, meaning we can change this one variable while holding everything else fixed, which is the standard requirement for controlled causal evaluation~\citep{ahmed2020causalworld,scholkopf2021toward}. The experts then merged candidates. 
For example, braking intensity and acceleration intensity look like opposite actions and a surface-level taxonomy would keep them apart, but they share the same underlying mechanism: a continuous scalar on the action side (how hard) whose value monotonically determines the outcome until it crosses a categorical boundary (e.g., stopping in time versus not). We therefore merge them into a single Force/Degree primitive. Applying this mechanism-based logic to all candidates and organizing the resulting primitives along the natural cause-and-effect structure of an embodied event yields a two-domain taxonomy: properties of the surrounding environment ($\mathcal{D}_{\text{env}}$) and parameters of the agent's action ($\mathcal{D}_{\text{int}}$), with three primitives in each.

\noindent\textbf{$\mathcal{D}_\text{env}$ --- Environment Domain.} External conditions that change what a given action produces. (1) \textit{Surface Friction} determines whether contact yields traction or slip: hard braking decelerates cleanly on dry asphalt but skids on ice. (2) \textit{Material \& Medium} sets the resistance and deformability of the bodies and the medium they move through, which shapes how force translates into motion, so that the same gripper closure holds a rigid block but crushes a sponge. (3) \textit{Obstacle Configuration} fixes what space the agent can actually move into, where a single misplaced cone flips the outcome from clean passage to collision.

\noindent\textbf{$\mathcal{D}_\text{int}$ --- Interaction Domain.} Parameters of the action itself, holding the environment fixed. (1) \textit{Spatial Alignment} is where the action is applied: small positional errors separate a clean merge from a sideswipe, or a grasp from a miss. (2) \textit{Force/Degree} is how hard or how far (force, speed, or angular extent), and scales the outcome continuously until it crosses a categorical boundary, such as stopping in time versus not, or holding versus dropping. (3) \textit{Temporal Sequencing} is when each sub-action fires relative to the others; closing the gripper before rather than after reaching the target turns the same motion into a grasp or a miss.

Table~\ref{tab:unified_taxonomy} instantiates each primitive in both domains. Each benchmark instance varies exactly one primitive under unanimous expert admission, so that a failure can be attributed to a specific reasoning deficit rather than to poor physics in general.

\subsection{Benchmark Construction}
\label{sec:construction}

\begin{figure}[tbp]
    \centering
    \includegraphics[width=1\textwidth]{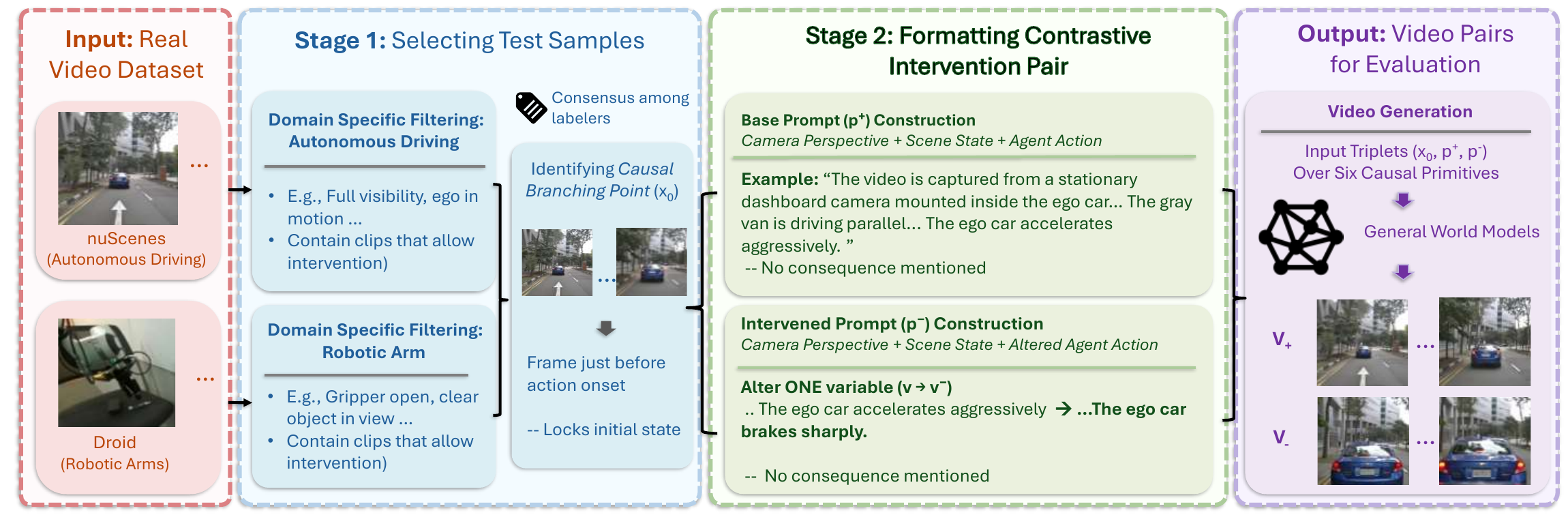}
    \caption{Benchmark construction pipeline. Stage 1 filters clips from nuScenes and DROID and extracts the causal branching point $x_0$ at action onset. Stage 2 writes a contrastive prompt pair $(p^+, p^-)$ that shares scene context and differs only in one physical variable, with the outcome withheld. Each triplet $(x_0, p^+, p^-)$ is fed to a video model to produce $(V^+, V^-)$.}
    \label{fig:pipeline}
\end{figure}

To solve the \textbf{How to test} challenge, we design each test instance as a controlled probe of the model's world-simulator capability from a
shared initial state, a change to one physical variable should drive the simulated rollout to diverge as physical law predicts. Two
requirements~\citep{pearl2009causality,scholkopf2021toward} support
the interpretability of this probe. The first is state fixation:
$\mathcal{V}_+$ and $\mathcal{V}_-$ should originate from the same
initial state, so that observed divergence can be attributed to $v$
rather than to differing initial conditions. The second is
single-variable change: only one key variable should differ between
the two prompts, so that divergence can be cleanly traced to a
specific mechanism. Existing video-generation
benchmarks~\citep{DBLP:conf/cvpr/HuangHYZS0Z0JCW24, bansal2024videophy,DBLP:conf/icml/MengLTLSZ0L025,li2025worldmodelbench}
satisfy neither and only test each video models against a single prompt in isolation. 
We meet both requirements with a two-stage construction
(Figure~\ref{fig:pipeline}): (1) selecting a real clip and extracting
a frame $x_0$ that, together with a fixed scene-state
description, anchors the initial state, and (2) formatting a paired
prompt $(p_+, p_-)$ with only semantic different in target
variables.

\noindent\textbf{Anchoring the state: $x_0$ at the causal branching point.} We anchor the initial state through a paired specification: a single image frame $x_0$ together with the \emph{camera perspective} and \emph{initial scene state} components of the prompt. The frame fixes what is visually observable (geometry, materials, agent positions, lighting), while the two prompt components supply information that a single frame cannot encode: how the camera will behave over time (e.g., \textit{``stationary dashboard camera''}) and dynamic context such as the ego vehicle's speed or its motion relative to other agents. This split mirrors how current video-generation world models are conditioned (image as visual anchor, text as the source of non-visual initial information), and the three together form the shared initial conditioning that we hold fixed across $\mathcal{V}_+$ and $\mathcal{V}_-$.

Frames are drawn from nuScenes~\citep{caesar2020nuscenes} (autonomous driving) and DROID~\citep{khazatsky2024droid} (robotic arm manipulation). Within these datasets, a clip is eligible only if it carries the \emph{physical affordances} that make a target primitive testable, such as the ego car driving side-by-side with another vehicle in an overtaking scenario for \emph{Surface Friction}, or an unoccluded gripper near a target object for manipulation primitives. Domain-specific visibility filters further remove parked ego vehicles, heavy occlusion, and pre/after-contact gripper states, so that any divergence between $\mathcal{V}_+$ and $\mathcal{V}_-$ remains visible against an uncluttered scene.

From each qualified clip we take the frame immediately preceding action onset as the visual component of $x_0$, and we refer to this moment as the \emph{causal branching point}. At this moment, $\mathcal{V}_+$ and $\mathcal{V}_-$ can start from an identical state (state fixation), and at the same time, the intervention $v$ has not yet materialized, so $v$ retains room to change the future (we call this \emph{physical affordance}). In the overtaking scenario, for instance, $x_0$ is the exact moment the ego car and an adjacent vehicle travel side-by-side at identical speeds, where the future trajectory depends entirely on the subsequent action. See Appendix~\ref{x0details} for more details.

\noindent\textbf{Isolating the variable: the contrastive prompt pair.} Since text is the primary control input that current video-generation world models accept, we implement the contrastive design in the prompt pair. With $x_0$ and the first two prompt components already fixed by the shared initial conditioning above, the only span left for editing is the agent action description. The intervention prompt $p_-$ is obtained by editing this span alone, so that $p_+$ and $p_-$ differ only in how they describe the target variable. For example, ``\textit{[stationary dashboard camera]\ldots a gray van drives parallel to the ego car at the same speed; after a short moment, the ego car accelerates aggressively.}'' Both prompts withhold the physical consequence: neither states whether the ego car ends up ahead or behind, so the model cannot retrieve the outcome directly from text priors and must instead simulate it from $x_0$ and $v$, which is the capability the benchmark probes.

The intervention prompt $p_-$ is obtained by editing only the
description of the target variable in the action span:
$p_- = p_+[\,v \leftarrow v_-\,]$; for the example above, the ego
action description flips from acceleration to hard deceleration.
Combined with the shared initial conditioning, this restricts the
prompt-level edit to a single, locatable change in the textual
specification, so that any divergence between $\mathcal{V}_+$ and
$\mathcal{V}_-$ should be most cleanly traceable to the model's
handling of $v$. The full prompt building process is in Appendix~\ref{app:prompt-protocol}.

\subsection{Evaluation Metrics}
\label{sec:evaluation}

To address \textbf{how to evaluate}, our metric must diagnose whether the generated pair $(\mathcal{V}_+, \mathcal{V}_-)$ genuinely reflects the expected causal effect of the intervention. Each video must individually be a valid generation: it should execute its action, obey physics, and maintain a stable scene. Otherwise, comparing $\mathcal{V}_+$ to $\mathcal{V}_-$ amounts to comparing two broken outputs. Prior video benchmarks~\citep{DBLP:conf/cvpr/HuangHYZS0Z0JCW24, bansal2024videophy, DBLP:conf/icml/MengLTLSZ0L025, li2025worldmodelbench} measure exactly this per-video property and stop there.
But individual plausibility is not sufficient: a model can produce two individually plausible videos that ignore $v$ entirely, generating identical or randomly different rollouts. Detecting this requires comparing $\mathcal{V}_+$ and $\mathcal{V}_-$ jointly. Prior work in vision--language benchmarks~\citep{thrush2022winoground, agrawal2018don} has shown that paired comparison is essential for exposing exactly this failure mode. However, existing evaluations of video world models have not adopted this design, even though downstream uses such as action-conditioned simulation~\citep{yang2023unisim} and model-based planning~\citep{hafner2023mastering} depend on this property.

\begin{table}[htbp]
    \centering
    \caption{\textbf{The APEO evaluation framework.} Each criterion is scored as a binary success ($1$) or fail ($0$); a fail corresponds to the negation of the listed pass criterion. Dimensions A, P, and E are evaluated under both single-video and paired modes, whereas O is paired-only.}
    \label{tab:apeo_main}
    \footnotesize
    \setlength{\tabcolsep}{4pt}
    \begin{tabular}{@{} l p{5.7cm} p{5.9cm} @{}}
        \toprule
        \textbf{Dimension} & \textbf{Single-Video Evaluation Criterion of Success ($s$)} & \textbf{Paired-Videos Evaluation Criterion of Success ($p$)} \\
        \midrule
        \textbf{A} \emph{Adherence}
        & $\mathcal{V}$ executes the action prescribed by $p$ on the intended 
        target entity.
        & The actions rendered in $\mathcal{V}_+$ and $\mathcal{V}_-$ are 
        visually distinguishable, and their difference is aligned with the 
        direction of the prompt-level controls. \\[3pt]
        \textbf{P} \emph{Physics}
        & Motion within $\mathcal{V}$ respects basic physical constraints 
        throughout (no teleportation, morphing, or ghost forces).
        & $\mathcal{V}_+$ and $\mathcal{V}_-$ share an aligned trajectory prior 
        intervention and bifurcate thereafter in a manner consistent 
        with physical law under the prescribed change. \\[3pt]
        \textbf{E} \emph{Environment}
        & Background, camera viewpoint, and non-target object permanence remain 
        stable throughout $\mathcal{V}$.
        & The backgrounds of $\mathcal{V}_+$ and $\mathcal{V}_-$ are near-identical; 
        observable cross-video difference is attributable to the intervened 
        action rather than to scene-level artefacts. \\[3pt]
        \textbf{O} \emph{Outcome}
        & \textit{Not applicable.} (Outcome is defined only relative to a counterfactual partner.)
        & $\mathcal{V}_+$ and $\mathcal{V}_-$ terminate in measurably distinct 
        final states, with the divergence aligned to the prediction of $v$ in 
        both direction and affected entities. \\
        \bottomrule
    \end{tabular}
\end{table}

\noindent\textbf{The APEO framework.} Adopting paired comparison as our evaluation design leaves one question open: what should we compare $\mathcal{V}_+$ and $\mathcal{V}_-$ on? We design a four-dimension framework whose dimensions jointly cover the conditions a pair must meet to reflect the causal effect of $v$: Adherence (the intervention itself was executed), Physics (the dynamics were physically valid), Environment (the context in video held still), and Outcome (the result differed in the predicted direction). The four dimensions are intended to be complementary, each focusing on a different aspect of how a pair can fail. Each one also corresponds to concrete downstream consumers of a video world model:
\textit{Adherence of Controls~(A)} matters for VLA policy
training~\citep{wang2025alpamayo, kim2024openvla, zitkovich2023rt} and
action-conditioned world models~\citep{yang2023unisim, bruce2024genie}:
if $\mathcal{V}_-$ silently performs a different action, the policy
learns a wrong action--outcome map. \textit{Physical Realism~(P)} matters for inverse dynamics
learning~\citep{baker2022video} and latent dynamics
modeling~\citep{hafner2023mastering}: physically impossible
intermediate frames corrupt the dynamics signal even when endpoints
look correct~\citep{DBLP:conf/icml/KangYL00W0F25}. \textit{Environmental
Consistency~(E)} matters for synthetic data
augmentation~\citep{wang2023gensim,hu2023gaia}: hallucinated
background changes induce shortcut
learning~\citep{geirhos2020shortcut}. \textit{Outcome Divergence~(O)}
matters for model-based planning and counterfactual
simulation~\citep{hafner2023mastering,yang2023unisim}: their value
depends on rolling out distinct futures under distinct actions.

Each APEO dimension except $O$ can be evaluated in two modes
(Table~\ref{tab:apeo_main}). The single-video checks ($A_s, P_s, E_s$)
ask whether each video alone executes the prompted action, obeys
physics, and holds its scene stable; these are necessary for
plausibility but silent on whether the pair diverges correctly. The
paired-video checks ($A_p, P_p, E_p, O_p$) enforce \emph{ceteris
paribus}: $A_p$ checks that $\mathcal{V}_+$ and $\mathcal{V}_-$
execute visually distinct actions; $P_p$ checks that their
trajectories overlap before the intervention and diverge plausibly
afterward; $E_p$ checks that their backgrounds match, so any visible
difference is attributable to $v$; and $O_p$ checks that the resulting
states differ in the direction $v$ predicts.

The two modes catch different failures. A model can pass every
single-video check yet fail $P_p$ and $O_p$ if both videos are
individually plausible but follow the same trajectory, which is the
contrastive bottleneck (Section~\ref{sec:results}) where the model ignores
$v$ entirely. Conversely, a pair can pass $O_p$ for the wrong reason,
diverging through hallucinated scene artifacts rather than through
$v$, a failure only $E_p$ catches. Reporting all these
dimensions therefore helps localize where a model's causal pipeline
fails.

\noindent \textbf{VLM-based evaluation with human verification.}
Manual annotation at this scale is infeasible, so we use a VLM judge (Gemini 3.1 Pro), 
as is now standard in video generation benchmarks~\citep{DBLP:conf/cvpr/HuangHYZS0Z0JCW24, bansal2024videophy, 
li2025worldmodelbench}, recent autonomous-driving evaluation pipelines~\citep{wang2025alpamayo}, 
and consistent with the broader LLM-as-a-Judge literature~\citep{zheng2023judging, liu2023g}. For each APEO check,
the judge receives $(\mathcal{V}_+, \mathcal{V}_-)$ along with
$(p_+, p_-)$ and answers a primitive-conditioned binary question
(e.g., ``Does $\mathcal{V}_-$ exhibit a shorter stopping distance
than $\mathcal{V}_+$?''). We use binary rather than Likert scoring
because graded ratings exhibit position, verbosity, and
central-tendency biases that compress
variance~\citep{wang2024large}; binary decisions avoid these while
still aggregating to fine-grained scores through dimension- and
instance-level averaging. We validate against 421 human-annotated samples: the judge 
agrees with human labels on 82.30\% of decisions averaged 
across dimensions, comparable to the inter-human agreement of 84.03\%.

\section{Experimental Results}
\label{sec:results}

We evaluate nine video generation models spanning both open-source and closed-source tiers.
The \textbf{open-source} models include HunyuanVideo-1.5~\citep{wu2025hunyuanvideo}, Cosmos-Predict2~\citep{ali2025world},
CogVideoX~\citep{yang2024cogvideox}, and Wan2.2~\citep{wan2025wan}.
The \textbf{closed-source} models include Grok~Imagine~\citep{xai2025grokimagine}, Veo~3.1~\citep{deepmind2025veo3},
Seedance~2.0~\cite{seedance2026seedance}, Seedance~1.5~\citep{seedance2025seedance}, and Kling~3.0~\citep{kling2025omni}.
All models receive the same initial frame $x_0$ and prompt pair $(p_+, p_-)$; no
model is fine-tuned or adapted for this benchmark.
Each test instance yields one video per prompt ($\mathcal{V}_+$ and $\mathcal{V}_-$),
and every result reported in this section is expressed as a pass rate.

\subsection{The Benchmark Leaderboard}
\label{sec:main_results}

Table~\ref{tab:main_results} reports the full APEO scorecard.
We separate single-video scores ($s$, defined as the mean of the scores assigned to $\mathcal{V}_+$ and
$\mathcal{V}_-$ when each is assessed independently) from paired scores ($p$, evaluated on the
joint pair under the ceteris-paribus constraint); \textbf{O} denotes a paired-only metric.
The rightmost columns report two aggregate scores: $s$\textbf{Avg} = mean($A_s$, $P_s$, $E_s$), which summarises single-video quality, and $p$\textbf{Avg} (our primary \textbf{APEO} metric) = mean($A_p$, $P_p$, $E_p$, $O_p$), which measures causal sensitivity under paired evaluation.

\definecolor{rank1}{HTML}{FFCCCC}
\definecolor{rank2}{HTML}{FFE0CC}
\definecolor{rank3}{HTML}{CCE0FF}
\newcommand{\first}[1]{\cellcolor{rank1}{#1}}
\newcommand{\second}[1]{\cellcolor{rank2}{#1}}
\newcommand{\third}[1]{\cellcolor{rank3}{#1}}

\begin{table*}[tbp]
    \centering
    \caption{\textbf{What-If World Benchmark Leaderboard.}
    All scores are pass rates (\%).
    Best results are in \colorbox{rank1}{red}, second best in \colorbox{rank2}{orange}, third best in \colorbox{rank3}{blue}.
    $^*$~indicates closed-source models.}
    \label{tab:main_results}
    \begin{tabular}{@{} l cc cc cc c cc c @{}}
        \toprule
        \multirow{2}{*}{\textbf{Model}} &
        \multicolumn{2}{c}{\textbf{A}} &
        \multicolumn{2}{c}{\textbf{P}} &
        \multicolumn{2}{c}{\textbf{E}} &
        \textbf{O} &
        \multicolumn{2}{c}{\textbf{APEO}} &
        \multirow{2}{*}{\textbf{Rk.}} \\
        \cmidrule(lr){2-3}\cmidrule(lr){4-5}\cmidrule(lr){6-7}\cmidrule(lr){9-10}
        & $s$ & $p$ & $s$ & $p$ & $s$ & $p$ & $p$ & $s$Avg & $p$Avg & \\
        \midrule
        CogVideoX1.5-5B  & 32.8 & 18.2 & 42.5 & 10.7 & 27.3 & 50.8 & 11.3 & 34.2 & 22.7 & 9 \\
        Wan2.2-5B            & 40.4 & 22.3 & 52.0 & 14.1 & 45.9 & 60.2 & 17.2 & 46.1 & 28.4 & 7 \\
        HunyuanVideo1.5-8.3B  & 42.6 & 20.4 & 64.4 & 12.2 & 48.0 & 64.9 & 10.7 & 51.7 & 27.0 & 8 \\
        Cosmos-Predict2-2B  & 41.4 & 25.1 & 55.2 & 18.8 & 55.8 & \second{73.0} & 14.7 & 50.8 & 32.9 & 5 \\
        \midrule
        Seedance 1.5$^*$  & 51.6 & 32.0 & 58.6 & 20.7 & 47.8 & 51.1 & 26.0 & 52.7 & 32.4 & 6 \\
        Kling 3.0$^*$     & 54.1 & 34.8 & 65.5 & 23.8 & 66.5 & \first{74.9} & 21.0 & 62.0 & 38.6 & 4 \\
        Seedance 2.0$^*$  & \third{63.0} & \third{41.7} & \first{74.0} & \third{28.8} & \first{71.5} & 64.9 & \third{32.0} & \first{69.5} & \third{41.8} & 3 \\
        Veo 3.1$^*$         & \second{66.8} & \second{48.0} & \third{69.7} & \second{40.1} & \second{70.1} & \third{70.5} & \second{44.8} & \third{68.9} & \second{50.9} & 2 \\
        Grok Imagine$^*$  & \first{69.2} & \first{49.2} & \second{70.5} & \first{42.6} & \third{67.2} & 69.7 & \first{45.1} & \second{69.0} & \first{51.7} & 1 \\
        \bottomrule
    \end{tabular}
\end{table*}

\begin{figure}
    \centering
    \includegraphics[width=1\linewidth]{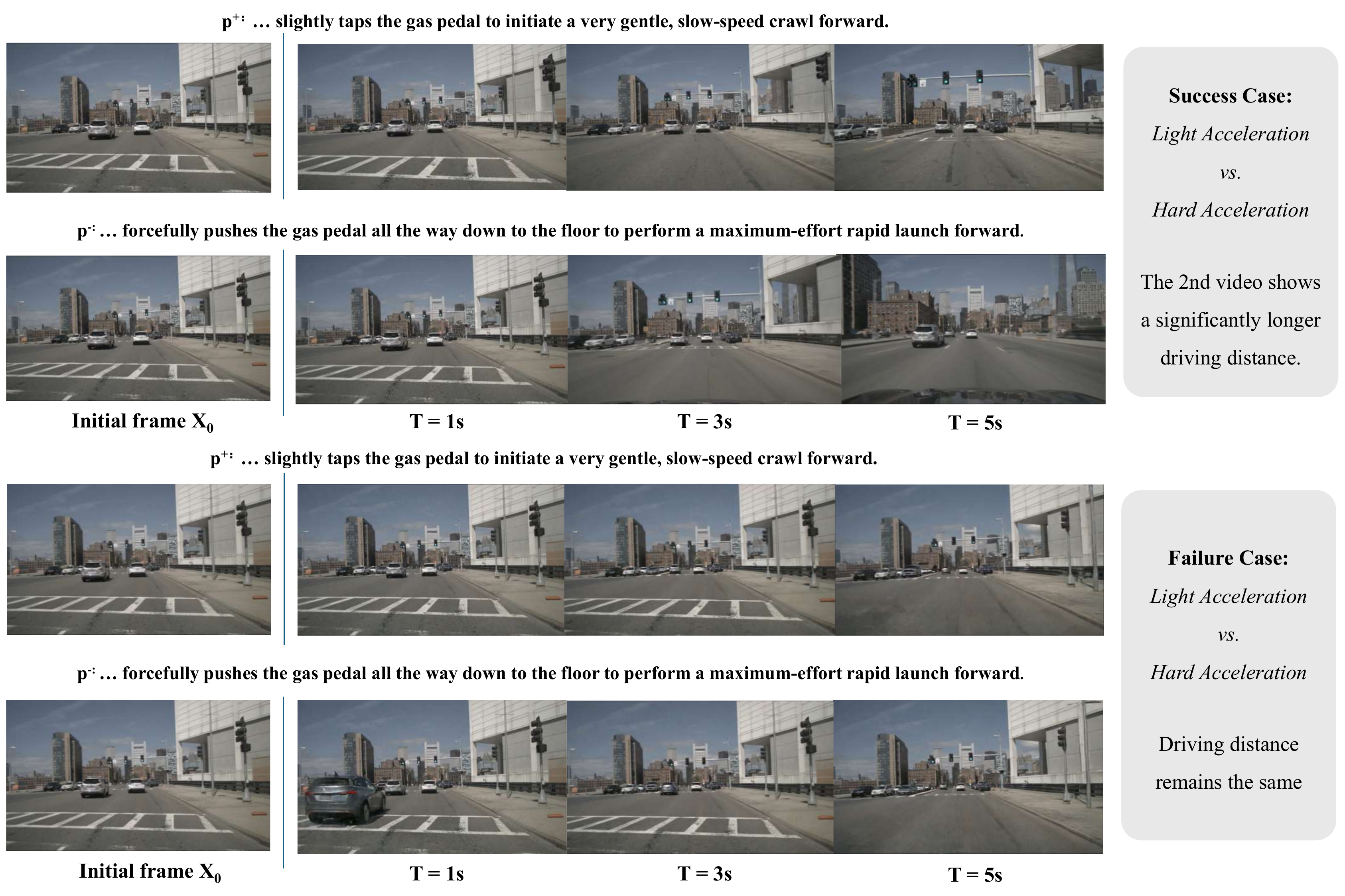}
    \caption{\textbf{Contrastive Bottleneck Example (AD).} Both panels
    start from the same initial frame $x_0$ and receive a Force/Degree
    intervention pair that differs only in the acceleration verb---$V_+$:
    ``a gentle, slow-speed crawl forward''; $V_-$: ``a maximum-effort rapid
    launch forward.'' Frames are sampled at $T=1, 3, 5$\,s. The difference in driving distance is the quantity that causal evaluation measures.}
    \label{fig:contrastive_bottleneck1}
\end{figure}

\begin{figure}
    \centering
    \includegraphics[width=1\linewidth]{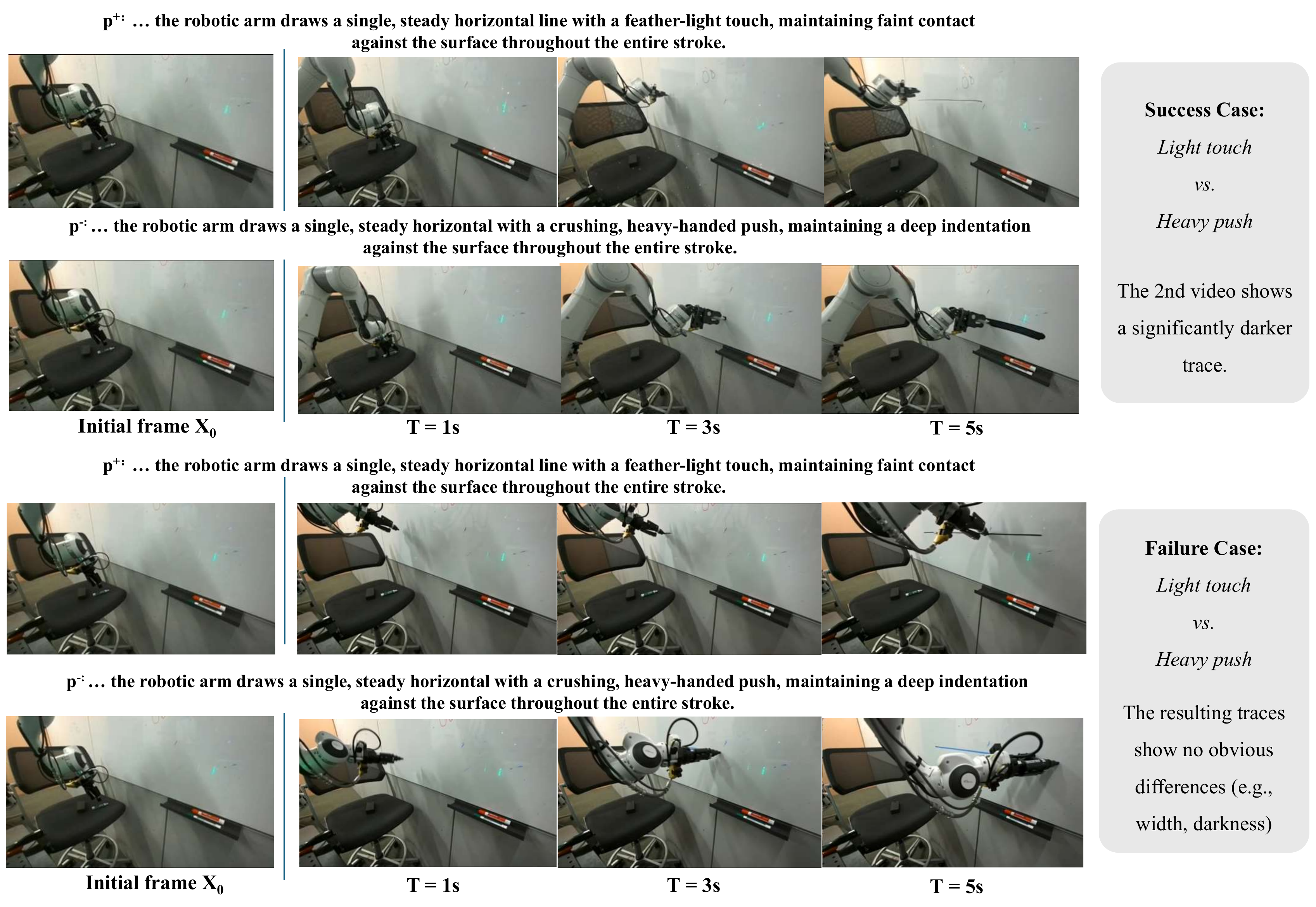}
    \caption{\textbf{Contrastive Bottleneck Example (Robotic Arms).} 
Both panels start from the same initial frame $x_0$ and receive a Force/Pressure intervention pair that differs only in contact intensity---$V^{+}$: ``a gentle, feather-light touch maintaining faint contact''; $V^{-}$: ``a heavy-handed push producing a deep indentation.'' Frames are sampled at $T = 1, 3, 5$\,s. The discriminability of the resulting contact trace (e.g., its darkness and width) is the property that causal evaluation measures.}
    \label{fig:contrastive_bottleneck2}
\end{figure}

\noindent \textbf{Observations.}
Three findings emerge.
\textbf{(1) Causal reasoning remains largely unsolved.}
The best-performing model, Grok~Imagine, achieves only 51.7\% APEO, indicating that it fails on nearly half of all causal tasks and there is still substantial room for improvement.
\textbf{(2) The closed-source tier leads consistently.}
Closed-source models attain an average APEO of 43.1\%, compared with 27.8\% for open-source models---a gap of 15.3\, percentage points (pp).
\textbf{(3) Single-video scores consistently overestimate capability.}
Every model attains a higher score on single-video metrics than on paired metrics for both Action and Physics.
For instance, HunyuanVideo-1.5 reaches $P_s = 64.4\%$ but only $P_p = 12.2\%$---a drop of 52.2\,pp.
We refer to this discrepancy as the \emph{contrastive bottleneck}: the generation of an individually plausible video does not translate into correct causal differentiation across a contrastive pair. See Figure~\ref{fig:contrastive_bottleneck1},~\ref{fig:contrastive_bottleneck2} and Appendix~\ref{app:qualitative} for qualitative examples.

\subsection{Why Paired Evaluation Matters: The Hidden Failure Stratum}
\label{sec:paired-matters}

The gap between single-video and paired scores observed in Table~\ref{tab:main_results} is not a mere calibration nuisance but a structural feature of how current models fail. We document the following consequences, which demonstrate that paired evaluation is not simply more demanding than single-video evaluation but rather \emph{measures something different}. 

\textbf{Single-video and paired evaluations produce different leaderboards.} Ranked by single-video quality $s_{\text{Avg}}$, the top three are Seedance 2.0 (69.5\%), Grok Imagine (69.0\%), and Veo~3.1 (68.9\%). Ranked by the primary paired metric $p_{\text{Avg}}$, however, the ordering shifts to Grok Imagine (51.7\%), Veo~3.1 (50.9\%), and Seedance 2.0 (41.8\%): the single-video winner falls to third, with a 27.7\,pp gap between its $s_{\text{Avg}}$ and $p_{\text{Avg}}$, the largest in the top tier. Conversely, Cosmos-Predict2 rises from 7th on $s_{\text{Avg}}$ to 5th on $p_{\text{Avg}}$. Identifying the best model therefore depends on which evaluation paradigm the community adopts. See Appendix~\ref{app:addition} for more results.

\textbf{Some failures are invisible to single-video metrics.}
Some failures can escape per-video inspection: both videos individually appear flawless (the prompted action is executed, the physics looks plausible, and the scene is stable), yet when compared side by side, they fail to diverge in the manner predicted by the underlying physical variable.
For example, a hard-brake video and a gentle-brake video may each exhibit smooth and realistic decelerations, yet the resulting stopping distances are indistinguishable.
Only a paired comparison reveals that the model has ignored the difference between the two inputs.
Among the 314 instances where every single-video check passes, paired physics still fails in 13.1\% of cases, and the overall outcome fails in 8.3\%.
These hidden failures pose the greatest risk for downstream applications such as counterfactual planning: a planner relying on such a model would receive two convincing yet causally identical futures, with no indication that the simulation failed to distinguish the two inputs.

\subsection{When World Model Fails: Per-Primitive Analysis}
\label{sec:anatomy}
We decompose the aggregate scores by the six physical primitives (See Appendix Tables~\ref{tab:prim_spatial_alignment}--\ref{tab:prim_obstacle_configuration}).
The per-primitive breakdown reveals that difficulty is not uniform: it maps onto interpretable factors that distinguish what current models can do from what they fundamentally cannot.

\textbf{Models appear to respond to visual cues more than latent physical state.}
Model performance varies sharply across the six primitives (by $O_p$): Force/Degree (40.4\%) $>$ Spatial Alignment (30.9\%) $>$ Material/Medium (23.9\%) $>$ Obstacle Configuration (15.5\%) $\approx$ Temporal Sequencing (15.3\%) $\approx$ Surface Friction (14.2\%).
This ordering does not seem to be explained by physical complexity, since friction is arguably no harder to simulate than force.
A more plausible factor is whether the intervening variable is visible in the initial frame.
Force/Degree manipulates acceleration, which tends to produce dramatic, training-familiar speed changes ($O_p = 75.4\%$ on the highest-scoring sub-task).
Surface Friction and Object Mass, by contrast, manipulate latent properties—a light cup and a heavy block look identical in $x_0$ ($O_p = 7.4\%$ on Object Mass, the lowest-scoring sub-task).

\textbf{Temporal Sequencing represents a universal challenge.}
Unlike other primitives where strong models separate from weak ones (Force/Degree spans 26--71\% APEO, a 45\,pp range), Temporal Sequencing shows compressed performance: even the best model (Seedance 2.0) reaches only $O_p = 25.6\%$, and the closed-source advantage nearly vanishes (only 5.9\,pp vs.\ 23.2\,pp on Force/Degree).
Consider an overtaking scenario where the ego car should accelerate \emph{before} vs.\ \emph{after} the front car speeds up: the two timings should yield very different final gaps, but a representative VLM judge observation reports that \textit{``the relative position between vehicles is only marginally different between the two timing conditions''}.
The prompted acceleration is executed in each video, but its timing is not anchored to the front car's state, so the two conditions collapse to nearly the same relative geometry.
This relational coupling appears unaddressed by current scaling and data strategies.

\section{Conclusion}
\label{sec:conclusion}
This work introduces What-If World, the first benchmark for contrastive intervention following in video world models. What-If World combines (1) a six-primitive taxonomy of physical intervention variables spanning autonomous driving and robotic manipulation, (2) 319 paired test instances anchored on real frames from nuScenes and DROID with only one variable changed per pair, and (3) the APEO framework, which evaluates Adherence, Physics, Environment, and Outcome in both single-video and paired settings. Benchmarking nine state-of-the-art video generation models reveals a contrastive bottleneck that prior benchmarks miss: per-video plausibility does not guarantee a correct response to controlled input changes. The strongest model achieves only 51.7\% 
on the paired metric, and performance concentrates on visually familiar interventions rather than generalizing across the taxonomy. By making input sensitivity directly observable at scale, What-If World provides a unified and reproducible framework for diagnosing where current generative video systems struggle as world models, and highlights an important capability gap for embodied applications such as action-conditioned simulation, model-based planning, and vision-language-action policy training.

\noindent \textbf{Limitation and Future Work.} What-If World measures causal capability rather than downstream utility. Each APEO dimension is anchored to a concrete downstream consumer (Section~\ref{sec:evaluation}), but empirically validating how these scores translate into deployed performance is left to future work. %
We also characterize failure modes phenomenologically without attributing them to specific architectural or training-data causes; doing so requires training-side controlled experiments that fall outside the scope of an evaluation benchmark.

\bibliographystyle{plainnat}
\bibliography{refs}

\appendix

\section{Other Definitions and Criteria}
\label{sec:appendix_definition}

\begin{table*}[htbp]
    \centering
    \caption{\textbf{The Unified Causal Taxonomy across Embodied Domains.} 
Six physical primitives organized into two domains (Environment and 
Interaction), three per domain, each instantiated in both Autonomous 
Driving and Robotic Manipulation. Citations indicate prior 
physical-reasoning benchmarks grounding each primitive.}
    \label{tab:unified_taxonomy}
    \footnotesize
    \renewcommand{\arraystretch}{1.25}
    \setlength{\tabcolsep}{6pt}
    \begin{tabularx}{\textwidth}{@{} l l X X @{}}
        \toprule
        \textbf{Domain} & \textbf{Physical Primitive} & \textbf{Autonomous Driving} & \textbf{Robotic Manipulation} \\
        \midrule
        \multirow{3}{*}{\shortstack[l]{\textbf{D1: Env.}\\\textit{(Environmental}\\\textit{Conditions)}}}
            & Surface properties~\cite{patel2022cripp,tung2023physion++}
            & Dry asphalt vs.\ muddy / icy road surface
            & Smooth (glass) vs.\ grippy (rubber mat) surface \\
        \addlinespace[2pt]
            & Material \& medium~\cite{gao2024physically,zheng2024contphy}
            & Clear, dry weather vs.\ rain / snow
            & Rigid object (wood) vs.\ deformable object (sponge) \\
        \addlinespace[2pt]
            & Obstacle config.~\cite{bakhtin2019phyre}
            & Clear path vs.\ static obstacle ahead (cone, pothole)
            & Clear workspace vs.\ obstructing object in grasp path \\
        \midrule
        \multirow{3}{*}{\shortstack[l]{\textbf{D2: Int.}\\\textit{(Action}\\\textit{Mechanics)}}}
            & Spatial alignment~\cite{girdhar2019cater}
            & Lateral lane offset (safe merge vs.\ collision)
            & Gripper closure position (grasp success vs.\ miss) \\
        \addlinespace[2pt]
            & Force / degree~\cite{ates2022craft,liu2024physgen}
            & Light vs.\ hard braking; light vs.\ heavy acceleration
            & Small vs.\ large rotation angle; gentle vs.\ forceful push \\
        \addlinespace[2pt]
            & Temporal sequencing~\cite{girdhar2019cater,grunde2021agqa}
            & Speed / timing relative to leading car or merge point
            & Align gripper before closing vs.\ premature closure \\
        \bottomrule
    \end{tabularx}
\end{table*}

\subsection{Benchmark Statistics}
\label{app:dataset_stats}
The benchmark comprises 319 test instances spanning six physical primitives across two embodied domains: Autonomous Driving (AD, 11 sub-categories) and Robotic Arm Manipulation (Rob, 7 sub-categories), for 18 sub-categories in total. Each instance is evaluated by all 9 video generation models (4 open-source, 5 closed-source) with two generations per instance, yielding 5{,}742 generated videos. Table~\ref{tab:dataset_stats} summarizes the composition.

\begin{table}[htbp]
    \centering
    \caption{\textbf{Dataset composition.} Number of unique scenes and total evaluated instances (scenes $\times$ 9 models) per sub-category. AD = Autonomous Driving; Rob = Robotic Manipulation.}
    \label{tab:dataset_stats}
    \small
    \renewcommand{\arraystretch}{1.15}
    \begin{tabular}{@{} l l l r r @{}}
        \toprule
        \textbf{Primitive} & \textbf{Sub-category} & \textbf{Domain} & \textbf{Scenes} & \textbf{Instances} \\
        \midrule
        \multirow{5}{*}{Spatial Alignment} & Long. Accel & AD & 18 & 162 \\
         & Long. Brake & AD & 18 & 162 \\
         & Lateral Avoid & AD & 17 & 153 \\
         & Release Coords & Rob & 20 & 180 \\
         & Grasp Align & Rob & 22 & 198 \\
        \midrule
        \multirow{3}{*}{Force / Degree} & Brake Intensity & AD & 18 & 162 \\
         & Accel Intensity & AD & 14 & 126 \\
         & Push/Yank & Rob & 20 & 180 \\
        \midrule
        \multirow{4}{*}{Temporal Sequencing} & Overtake Hard & AD & 30 & 270 \\
         & Overtake Accel & AD & 15 & 135 \\
         & Overtake Brake & AD & 15 & 135 \\
         & Gripper Timing & Rob & 22 & 198 \\
        \midrule
        \multirow{2}{*}{Surface Friction} & Road Friction & AD & 15 & 135 \\
         & Table Friction & Rob & 15 & 135 \\
        \midrule
        \multirow{2}{*}{Material / Medium} & Road Medium & AD & 15 & 135 \\
         & Deformability & Rob & 15 & 135 \\
        \midrule
        \multirow{2}{*}{Obstacle Configuration} & Scene Layout & AD & 15 & 135 \\
         & Object Mass & Rob & 15 & 135 \\
        \midrule
        \textbf{Total} & & & \textbf{319} & \textbf{2871} \\
        \bottomrule
    \end{tabular}
\end{table}

\subsection{Compute Resources and Implementation Details}

\paragraph{Hardware.} All open-source video generation was performed on a single workstation running Ubuntu 22.04 with 4$\times$ NVIDIA RTX A6000 GPUs (48\,GB VRAM each). Each generation job was scheduled to a single GPU; we did not use tensor- or pipeline-parallel inference within a model.

\paragraph{Inference settings.} For every open-source model we used the publicly released default inference configuration---sampler, number of denoising steps, classifier-free guidance scale, and the model's default precision. No model was fine-tuned, distilled, or adapted to the benchmark, and no per-prompt hyperparameter tuning was performed. This isolates our paired-evaluation methodology from any model-side optimisation.

\paragraph{Per-video runtime.} Generating a single video on one A6000 takes 20--40 minutes, depending on the model, driven primarily by spatial resolution, frame count, and number of denoising steps. With four open-source models each producing 638 videos (319 pairs $\times$ 2), the total open-source generation budget is approximately 1{,}300 GPU-hours, completed in roughly two weeks of wall-clock time with the four GPUs running in parallel.

\paragraph{API-accessed components.} The five closed-source generation 
models (Grok Imagine, Veo 3.1, Seedance 2.0, Seedance 1.5, Kling 3.0) 
were accessed through hosted APIs via kie.ai~\cite{kieai_platform}, and 
each pair $(\mathcal{V}_+, \mathcal{V}_-)$ was scored by Gemini~3.1~Pro 
as a VLM judge over the four APEO dimensions, also via API. We therefore 
consumed no local GPU resources for either component, and end-to-end 
latency was governed by vendor-side queueing.

\subsection{Taxonomy Construction Protocol}
\label{app:taxonomy-protocol}

This section provides the full construction trace for the six-primitive taxonomy introduced in Section~\ref{sec:taxonomy}, following the thematic-analysis framework~\cite{braun2006using} adapted for the construction of a discrete causal-variable space.

\paragraph{Stage 1: Random Video Sampling.}
We drew 1{,}000 candidate videos by simple random sampling: 500 driving clips from nuScenes~\cite{caesar2020nuscenes} and 500 manipulation episodes from DROID~\cite{khazatsky2024droid}. Sampling used a fixed seed for reproducibility. We did not pre-stratify by scene type or task category, but the resulting pool spans the full diversity natively present in the two source datasets.

\noindent\textbf{Stage 2: Independent Open Coding.}
Two domain experts (one in autonomous driving, one in manipulation) independently reviewed their respective 500-video subset. For each clip, they recorded all physical variables whose perturbation would plausibly change the visible outcome. The AD expert proposed 18 candidate variables; the manipulation expert proposed 14, yielding 32 initial candidates across the two domains.

\noindent\textbf{Stage 3: Joint Adjudication and Merging.}
The two experts then convened to (a) apply the three admission criteria stated in Section~\ref{sec:taxonomy}---physically fundamental, embodiment-shared, operationally isolable---and (b) merge candidates that shared the same underlying causal mechanism. Two candidates were merged iff they (i) acted through the same physical mechanism on the outcome, and (ii) admitted a unified contrastive-intervention template. Of the 32 initial candidates, 14 were dropped during adjudication: some were rejected outright for failing an admission criterion, and others were absorbed into a broader sub-category as redundant within-domain duplicates. The remaining 18 sub-categories were then organized into six unified primitives.

A representative rejection is \textit{time of day}, along with the closely related \textit{ambient lighting}. We excluded both for failing criterion~(i), since they are perceptual variables that change how the scene is rendered rather than physical mechanisms governing how force translates into motion, and they cannot be cleanly isolated from co-varying properties such as shadows and contrast.

Representative within-domain merges include:
\begin{itemize}
\item In AD, \textit{left-lane shift} and \textit{right-lane shift} were merged into the direction-agnostic sub-category \textit{Lateral lane offset}, since both act through the same lateral-misalignment mechanism with sign-symmetric outcomes.
\item In AD, four overtaking-related variants, \textit{side passing}, \textit{overtake-then-merge}, \textit{decelerate to yield}, and \textit{decelerate-then-merge}, were collapsed into the unified sub-category \textit{Timing of overtaking vs.\ yielding}, since all four reduce to controlling \emph{when} the ego vehicle initiates its longitudinal action relative to a parallel-traveling neighbor; the optional follow-up lane change does not introduce a new causal variable.
\item In manipulation, \textit{top-down grasp} and \textit{side grasp} were merged into \textit{3D grasp alignment}, since both act through the same gripper--object spatial-alignment mechanism and differ only in approach direction.
\item In manipulation, \textit{object weight} and \textit{object density} were merged into \textit{Object mass}, since the variable that ultimately governs the contact dynamics is the integrated mass rather than its decomposition into volume and density.
\end{itemize}

Cross-domain unification then proceeded at the primitive level. For instance, braking intensity in AD and gripper push intensity in manipulation were placed under the same \textit{Force/Degree} primitive despite acting in different embodiments, because they share the same mechanism: a continuous scalar on the action side that monotonically scales the outcome.

\noindent\textbf{Stage 4: Reliability via Unanimous Consensus.}
Because the two experts independently coded \emph{disjoint} domain-specific subsets at Stage~2, we enforce a strict validity criterion at Stage~3: every retained sub-category and final primitive grouping was the product of \emph{unanimous agreement} between the two domain experts. Disagreements during adjudication were resolved by joint discussion until consensus, and no category was admitted by majority vote or external arbitration.

At the implementation stage of benchmark construction (Section~\ref{sec:construction}), some sub-categories were further split for convenience of test-instance authoring (e.g., \textit{Longitudinal relative distance control} was instantiated separately as longitudinal acceleration and longitudinal braking variants), yielding the 18 sub-categories used to organize the 319 test instances in Table~\ref{tab:dataset_stats}.

\subsection{\texorpdfstring{$x_0$}{x0} Selection Protocol}
\label{x0details}
This section provides the construction trace for the 319 visual anchors $x_0$ used in the benchmark. The protocol implements the two requirements stated in the main text---that $x_0$ lie at the causal branching point (the frame immediately preceding action onset, supporting both state fixation and causal affordance), and that the surrounding clip carry the physical affordances required by the target sub-category---through a three-stage pipeline combining automated screening with human consensus.

\noindent\textbf{Stage 1: VLM-based candidate selection.} For each of the 18 sub-categories defined in Section~\ref{sec:taxonomy}, we drafted a sub-category-specific eligibility specification listing (i) the physical affordances required for the target intervention to be testable (e.g., for \emph{Longitudinal Acceleration}: ego vehicle stationary or in slow motion with a clear lane ahead; for \emph{Grasp Alignment}: gripper open, target object unoccluded and within reachable workspace), (ii) the visibility requirements (agent and target fully visible, no heavy occlusion or motion blur, dashboard camera not parked, pre-contact gripper state in manipulation), and (iii) the definition of action onset for the intervention. We then prompted Gemini 3.1 Pro~\cite{team2023gemini} to scan candidate clips from nuScenes~\cite{caesar2020nuscenes} and DROID~\cite{khazatsky2024droid} against these specifications. For each eligible clip, the VLM additionally returned a candidate timestamp identifying the frame immediately preceding action onset, which we extract as the candidate $x_0$. Ineligible clips (e.g., missing affordances, insufficient visibility, or with the action already underway) were filtered out and not forwarded to human review.

\noindent\textbf{Stage 2: Independent human review.} Two researchers then independently reviewed every Stage~1 candidate (clip plus VLM-suggested $x_0$) and rendered a binary keep/reject decision. A candidate was rejected if any of the following held: (i) \emph{visibility}: occlusion, motion blur, or lighting obscures the target agent or object; (ii) \emph{ambiguous spatial relationships}: the frame does not unambiguously specify the geometry between agent and target (e.g., lateral offset between two vehicles, gripper--object alignment), so $\mathcal{V}_+$ and $\mathcal{V}_-$ would not share a well-defined initial geometry; (iii) \emph{state-fixation violation}: the candidate frame is mid-action rather than pre-action, so $v$ has already begun to materialize; or (iv) \emph{affordance mismatch}: the frame does not actually support the intended intervention despite Stage~1 admission. Visibility and ambiguous spatial relationships together accounted for the majority of rejections.

\noindent\textbf{Stage 3: Unanimous consensus.} Only candidates retained by \emph{both} reviewers were admitted to the benchmark; candidates kept by only one reviewer were discarded without further adjudication. This rule trades quantity for cleanliness: a frame whose admissibility is contested is precisely the kind whose initial state cannot be guaranteed fixed across $\mathcal{V}_+$ and $\mathcal{V}_-$, and admitting it would weaken the state-fixation guarantee underlying the paired-evaluation protocol.

\noindent\textbf{Environmental isolation via generative augmentation.}
For Environment-domain primitives, the real frame $x_0$ may contradict 
$p_-$ (e.g., a sunny road paired with an ``icy surface'' prompt). In such 
cases, we use Nano Banana (Gemini 3 Pro Image)~\citep{team2023gemini} to edit $x_0$, 
altering \emph{only} the target environmental condition (e.g., rendering 
snow, adding a traffic cone) while preserving all other scene elements. 
This ensures that $x_0$ aligns with both prompts and that the intervention 
isolates a single environmental variable.

\noindent\textbf{Per-sub-category sampling band.} We targeted 15--25 retained instances per sub-category, drawing additional Stage~1 candidates when a sub-category fell below 15 after Stage~3 attrition. The final counts in Table~\ref{tab:dataset_stats} reflect this band.

\subsection{Prompt Authoring Protocol}
\label{app:prompt-protocol}

All 319 contrastive pairs were authored under a three-stage human consensus protocol that follows the construction requirements specified in Section~\ref{sec:construction}: state fixation, single-variable change, and outcome withholding.

\noindent\textbf{Stage 1: Independent drafting}. Two researchers independently drafted $(p_+, p_-)$ pairs for each candidate clip, applying the three-part template introduced in Section~\ref{sec:construction} (camera perspective, initial scene state, agent action) and intervening only within the action span.

\noindent \textbf{Stage 2: Cross-review}. A third researcher reviewed every drafted pair against the Section~\ref{sec:construction} criteria, flagging any pair whose action description was inaccurate, inconsistent with the visible content of $x_0$, or potentially leaking the outcome.

\noindent \textbf{Stage 3: Unanimous adjudication}. All three researchers jointly discussed the flagged pairs, and only pairs receiving unanimous approval were retained. Approximately 37\% of initial drafts were rejected at this stage. The two dominant rejection reasons were (i) the action description did not faithfully match the affordances visible in $x_0$, and (ii) the wording was insufficiently specific to make the intended intervention unambiguous. Both reasons directly threaten the state-fixation guarantee, since if the prompt does not match what $x_0$ shows, $\mathcal{V}_+$ and $\mathcal{V}_-$ no longer originate from the same effective initial condition. We therefore treated these as hard rejections rather than candidates for revision.

Surviving pairs were then normalized to a uniform stylistic format (third-person present tense, fixed action-span opener) so that format unification did not reintroduce cross-pair content variance. As a final independent check on the single-variable-change requirement of Section~\ref{sec:construction}, every retained pair was passed through Gemini~3.1~Pro acting as an automated prompt-pair evaluator. For each pair, Gemini received $p_+$, $p_-$, and the name of the target causal variable $v$ (for example, ``braking intensity'' or ``surface friction''), and was instructed to judge whether the two prompts describe the same underlying scene, camera setup, objects, and initial conditions, and differ in exactly one aspect, namely the description of $v$. The instruction explicitly admitted minor wording variation that preserves meaning (synonyms, paraphrasing, equivalent grammatical constructions) as acceptable, since such variation does not introduce additional causal factors capable of affecting the generated video; only differences that would constitute an additional controlled variable were treated as violations. Gemini returned a binary judgment, together with, for any flagged pair, a list of the additional aspects in which the two prompts diverged. Flagged pairs were manually re-reviewed and either revised or rejected before being added to the released benchmark.

We use Gemini for this check despite its role as the video judge because the verification operates on text only and shares no plausible failure mode with the video-assessment task. The human three-way consensus remains the primary evidence of pair validity, with Gemini providing an additional independent layer. After the full pipeline (independent drafting, cross-review, unanimous adjudication, format normalization, and Gemini verification with manual revision of flagged cases), a final Gemini sweep over the released benchmark yielded a 95.5\% pass rate on the single-variable-change check, with the residual failures distributed evenly across primitives and consistent with random labeling noise rather than systematic protocol violations.

\section{Representative Failure Cases -- Qualitative Results}
\label{app:qualitative}

\begin{figure}[tbp]
    \centering
    \includegraphics[width=1\linewidth]{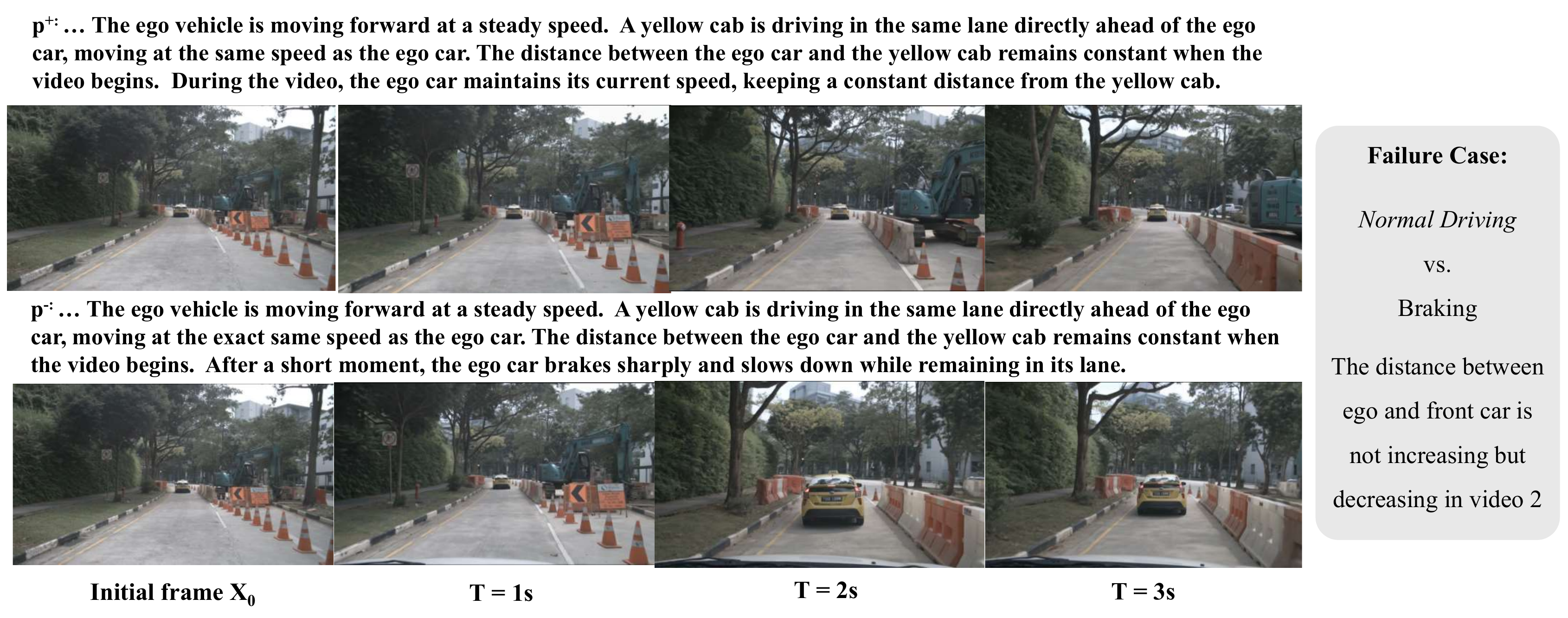}
    \caption{\textbf{Causal inversion (AD, Spatial Alignment).} The ego car is supposed to brake in $\mathcal{V}_-$, which should make the gap to the front car \emph{increase}. Instead, the gap \emph{decreases}---the outcome is reversed. $\mathcal{V}_+$ (constant speed) looks correct on its own, so only paired evaluation catches this failure.}
    \label{fig:fail_causal_inversion}
\end{figure}

\begin{figure}[tbp]
    \centering
    \includegraphics[width=1\linewidth]{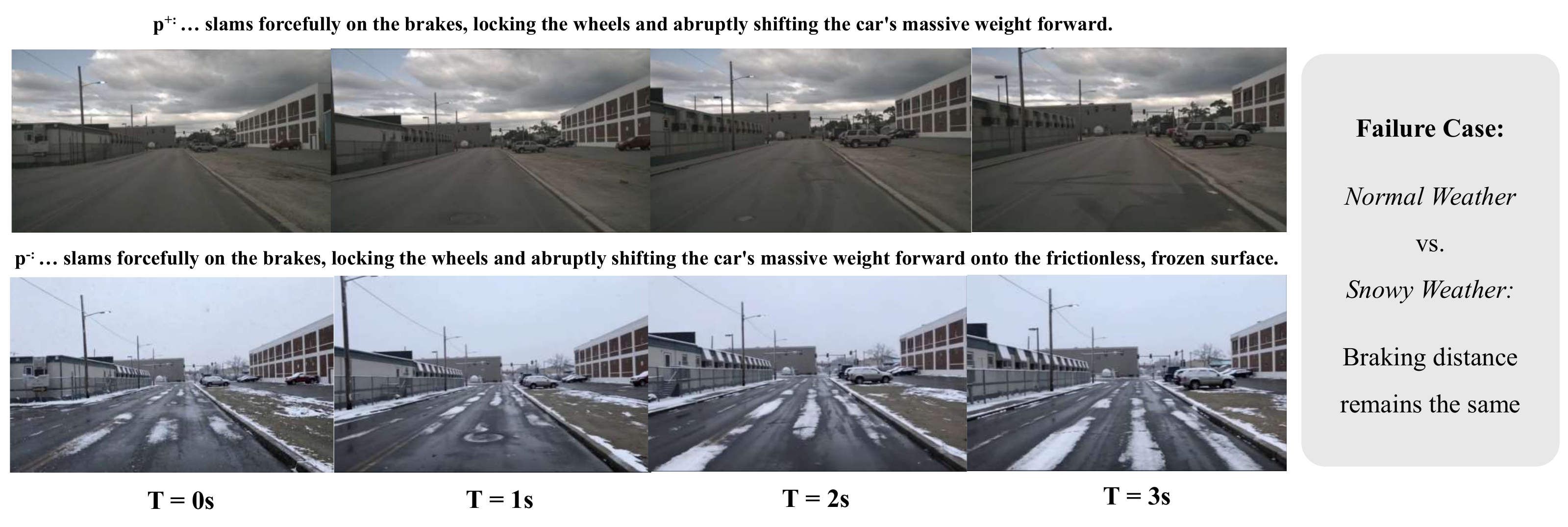}
    \caption{\textbf{Mode collapse under friction change (AD, Surface Friction).} Both videos show hard braking, but $p_-$ specifies a frozen, frictionless surface. The model renders snow correctly but produces the same braking distance in both conditions---the surface property has no physical effect on the outcome.}
    \label{fig:fail_friction}
\end{figure}

\begin{figure}[tbp]
    \centering
    \includegraphics[width=1\linewidth]{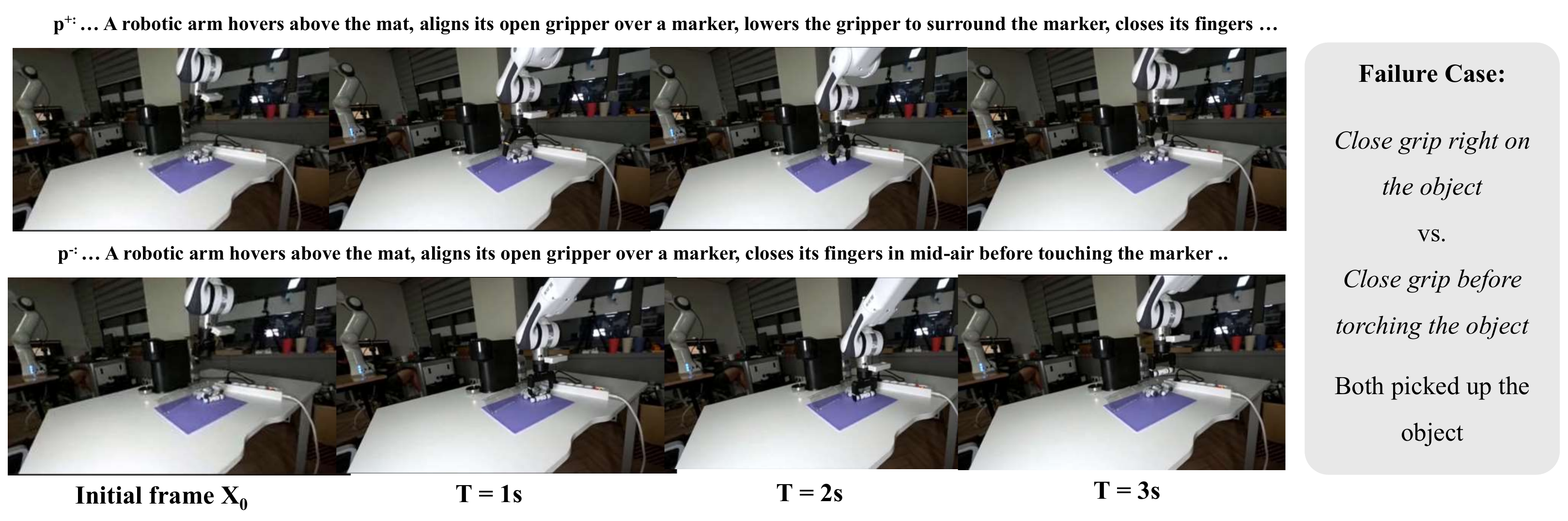}
    \caption{\textbf{Temporal reasoning failure (Robotic, Temporal Sequencing).} $p_+$ closes the gripper after lowering it onto the marker; $p_-$ closes the gripper in mid-air before touching the marker. The premature closure in $p_-$ should cause a miss, but the model picks up the object in both videos---it does not distinguish the two action sequences.}
    \label{fig:fail_temporal}
\end{figure}

\begin{figure}[tbp]
    \centering
    \includegraphics[width=1\linewidth]{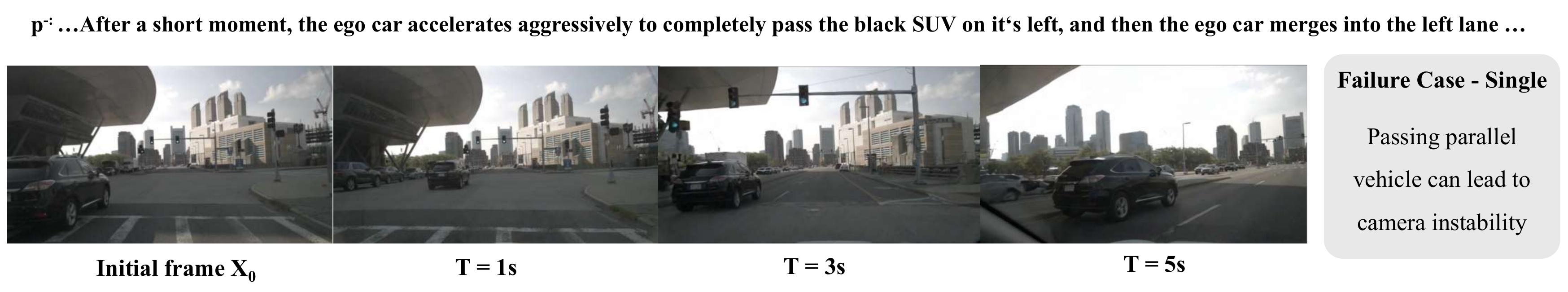}
    \caption{\textbf{Single-video camera hallucination (AD, $E_s$ failure).} The ego car accelerates to pass a black SUV and merge left. The model executes the action, but the camera viewpoint drifts severely over time, breaking the stationary dashboard-camera assumption. By $T=5$\,s the perspective no longer matches the original viewpoint.}
    \label{fig:fail_camera}
\end{figure}

The tables in the previous sections show \emph{how often} models fail; this section shows \emph{what those failures actually look like}. We select four representative examples, each illustrating a different failure mode from our taxonomy (Section~\ref{sec:failure_analysis}). For each case, we show the initial frame $x_0$, the two prompts ($p_+$ and $p_-$), and sampled frames from the generated videos, so the reader can see the problem directly.

\paragraph{Causal inversion in driving (Figure~\ref{fig:fail_causal_inversion}).}
This is a Spatial Alignment test from Autonomous Driving. The two prompts share the same setup (i.e., an ego car following a yellow cab at the same speed) and differ only in whether the ego car continues at constant speed ($p_+$) or brakes sharply ($p_-$). In a correct generation, the gap between the two cars should \emph{increase} in $\mathcal{V}_-$ (because the ego car slows down while the cab keeps going). Instead, the model produces a $\mathcal{V}_-$ where the gap \emph{decreases}---the ego car appears to get closer to the cab despite supposedly braking. This is a textbook causal inversion: the outcome moves in the opposite direction from what the physics predicts. Meanwhile, $\mathcal{V}_+$ looks reasonable on its own. A single-video evaluation would not catch this; only comparing the pair reveals that the causal relationship is reversed.

\paragraph{Mode collapse under surface friction change (Figure~\ref{fig:fail_friction}).}
This is a Surface Friction test. Both prompts describe the same hard-braking action, but $p_+$ takes place on a normal dry road while $p_-$ takes place on a frictionless, frozen surface. On ice, the car should slide much farther before stopping. However, the model produces nearly identical braking behavior in both conditions---the car decelerates at roughly the same rate and covers roughly the same distance. The model successfully renders the visual difference between the two environments (dry road vs.\ snowy road), but the changed surface has no effect on the physics. This is mode collapse at the outcome level: the model treats friction as a visual attribute (snow on the ground) rather than a physical property that changes how braking works.

\paragraph{Temporal reasoning failure in manipulation (Figure~\ref{fig:fail_temporal}).}
This is a Temporal Sequencing test from Robotic Manipulation. The two prompts describe the same gripper-over-marker setup, but $p_+$ instructs the arm to lower the gripper onto the marker \emph{before} closing, while $p_-$ instructs it to close its fingers in mid-air \emph{before} touching the marker. In reality, closing the gripper too early means it cannot grasp anything---it should miss. But the model generates a successful grasp in both videos: the marker gets picked up regardless of timing. The model has learned the visual pattern ``gripper near object $\rightarrow$ object gets picked up'' without understanding that the \emph{sequence} of close-then-lower vs.\ lower-then-close determines whether contact is made. This failure is representative of the Temporal Sequencing primitive more broadly, where even the best models struggle (Section~\ref{sec:anatomy}).

\paragraph{Single-video camera hallucination (Figure~\ref{fig:fail_camera}).}
Unlike the previous three examples, this is a single-video failure ($E_s$) rather than a paired one. The prompt asks the ego car to accelerate aggressively, pass a black SUV, and merge left. The model executes the overtaking action, but the camera viewpoint drifts dramatically over the course of the video---by $T = 5$\,s, the perspective has shifted so far that the scene looks like it was filmed from a different vehicle entirely. This kind of camera instability is a common $E_s$ failure in driving scenarios that involve lateral motion (lane changes, merges), where the model apparently struggles to keep the dashboard-camera viewpoint fixed while generating large lateral displacements.

\section{Additional Results and Analysis}
\label{app:addition}

\subsection{Per-primitive Analysis Results}

Tables~\ref{tab:prim_spatial_alignment}--\ref{tab:prim_obstacle_configuration} break down each model's APEO scores by physical primitive. We walk through them in order of decreasing average $O_p$, from the easiest primitive to the hardest.

\paragraph{Spatial Alignment (Table~\ref{tab:prim_spatial_alignment}).}
This primitive tests whether models can generate different outcomes when the \emph{position} of an action changes---for example, whether the ego car accelerates in the correct lane, or whether the gripper closes at the right spot.
Veo~3.1 leads with 60.8\% APEO, closely followed by Grok~Imagine at 54.5\%.
The $P_s$-to-$P_p$ drop is moderate here compared to other primitives: the top models retain roughly two-thirds of their single-video physics scores when evaluated as pairs, suggesting that spatial differences are among the easier causal signals for models to produce distinctly.
Open-source models lag substantially, with CogVideoX at just 26.3\%.

\begin{table}[tbp]
    \centering
    \caption{\textbf{Spatial Alignment} — per-model results (\%). Columns as in Table~\ref{tab:main_results}.}
    \label{tab:prim_spatial_alignment}
    \renewcommand{\arraystretch}{1.15}
    \begin{tabular}{@{} l cccccccc @{}}
        \toprule
        \textbf{Model} & $A_s$ & $A_p$ & $P_s$ & $P_p$ & $E_s$ & $E_p$ & $O_p$ & \textbf{APEO} \\
        \midrule
        CogVideoX1.5 & 34.2 & 22.1 & 37.4 & 22.1 & 24.2 & 44.2 & 16.8 & 26.3 \\
        Wan2.2 & 37.9 & 25.3 & 49.5 & 21.1 & 47.9 & 66.3 & 23.2 & 33.9 \\
        HunyuanVideo1.5 & 43.7 & 16.8 & 70.0 & 26.3 & 51.6 & 62.1 & 13.7 & 29.7 \\
        Cosmos-Predict2 & 38.9 & 27.4 & 67.9 & 32.6 & 59.5 & \second{74.7} & 17.9 & 38.2 \\
        \midrule
        Seedance 1.5$^*$ & 53.2 & 33.7 & 57.9 & 25.3 & 44.7 & 42.1 & 28.4 & 32.4 \\
        Kling 3.0$^*$ & 54.2 & 33.7 & 72.6 & 38.9 & 66.8 & \first{77.9} & 26.3 & 44.2 \\
        Seedance 2.0$^*$ & \second{66.3} & \third{45.3} & \first{80.0} & \third{43.2} & \third{69.5} & 63.2 & \third{42.1} & \third{48.4} \\
        Veo 3.1$^*$ & \first{69.5} & \first{53.7} & \second{77.4} & \first{57.9} & \first{72.6} & \third{71.6} & \first{60.0} & \first{60.8} \\
        Grok Imagine$^*$ & \third{63.7} & \second{50.5} & \third{76.3} & \second{52.6} & \second{72.1} & 65.3 & \second{49.5} & \second{54.5} \\
        \bottomrule
    \end{tabular}
\end{table}

\paragraph{Force / Degree (Table~\ref{tab:prim_forcedegree}).}
This is the easiest primitive. It asks whether models can show different \emph{magnitudes} of the same action---gentle vs.\ hard braking, or a light push vs.\ a forceful yank.
A possible explanation for the high score is that differences in acceleration or force produce visually obvious motion changes (the car moves much faster or much slower), and such scenarios are well represented in training data.
This is also where the closed-source vs.\ open-source gap is largest: the top three closed-source models all exceed 58\%, while no open-source model breaks 42\%.

\begin{table}[tbp]
    \centering
    \caption{\textbf{Force / Degree} — per-model results (\%). Columns as in Table~\ref{tab:main_results}.}
    \label{tab:prim_forcedegree}
    \renewcommand{\arraystretch}{1.15}
    \begin{tabular}{@{} l cccccccc @{}}
        \toprule
        \textbf{Model} & $A_s$ & $A_p$ & $P_s$ & $P_p$ & $E_s$ & $E_p$ & $O_p$ & \textbf{APEO} \\
        \midrule
        CogVideoX1.5 & 39.4 & 19.2 & 56.7 & 13.5 & 27.9 & 55.8 & 15.4 & 26.0 \\
        Wan2.2 & 37.5 & 19.2 & 47.1 & 17.3 & 38.5 & 57.7 & 23.1 & 29.3 \\
        HunyuanVideo1.5 & 45.2 & 19.2 & 64.4 & 11.5 & 46.2 & \first{76.9} & 19.2 & 31.7 \\
        Cosmos-Predict2 & 53.8 & 30.8 & 67.3 & 26.9 & 54.8 & 75.0 & 34.6 & 41.8 \\
        \midrule
        Seedance 1.5$^*$ & 59.6 & 42.3 & 57.7 & 40.4 & 46.2 & 55.8 & \third{53.8} & 48.1 \\
        Kling 3.0$^*$ & 45.2 & 28.8 & \second{75.0} & 25.0 & \second{65.4} & \first{76.9} & 28.8 & 39.9 \\
        Seedance 2.0$^*$ & \second{71.2} & \second{57.7} & \first{82.7} & \third{46.2} & \first{71.2} & \first{76.9} & \third{53.8} & \third{58.7} \\
        Veo 3.1$^*$ & \third{65.4} & \second{57.7} & 65.4 & \second{50.0} & \third{60.6} & 71.2 & \second{59.6} & \second{59.6} \\
        Grok Imagine$^*$ & \first{77.9} & \first{69.2} & \third{74.0} & \first{63.5} & 55.8 & 75.0 & \first{75.0} & \first{70.7} \\
        \bottomrule
    \end{tabular}
\end{table}

\paragraph{Temporal Sequencing (Table~\ref{tab:prim_temporal_sequencing}).}
This primitive is about \emph{when} sub-actions happen relative to each other---for example, whether the ego car accelerates before or after a gap opens up, or whether the gripper closes before or after touching the object.
It is one of the hardest primitives, and notably, the performance range across models is compressed: even the best model (Veo~3.1) only reaches 40.5\%, and the gap between Veo~3.1 and the worst model (Seedance~1.5, 20.4\%) is just 20\,pp---much smaller than the 45\,pp range on Force/Degree.
The $P_p$ scores are strikingly low across the board (the best is 19.5\%), meaning almost no model can produce trajectory pairs that diverge at the right moment.
This suggests that coordinating the timing of one agent's action relative to another agent's state is a fundamentally different challenge from simply generating different action intensities.

\begin{table}[tbp]
    \centering
    \caption{\textbf{Temporal Sequencing} — per-model results (\%). Columns as in Table~\ref{tab:main_results}.}
    \label{tab:prim_temporal_sequencing}
    \renewcommand{\arraystretch}{1.15}
    \begin{tabular}{@{} l cccccccc @{}}
        \toprule
        \textbf{Model} & $A_s$ & $A_p$ & $P_s$ & $P_p$ & $E_s$ & $E_p$ & $O_p$ & \textbf{APEO} \\
        \midrule
        CogVideoX1.5 & 30.5 & 11.0 & 62.8 & 6.1 & 35.4 & 73.2 & 12.2 & 25.6 \\
        Wan2.2 & 47.0 & 14.6 & 70.7 & 11.0 & 50.0 & 70.7 & \third{19.5} & 29.0 \\
        HunyuanVideo1.5 & 33.5 & 8.5 & \second{75.6} & 2.4 & 59.8 & 78.0 & 2.4 & 22.9 \\
        Cosmos-Predict2 & 32.3 & 14.6 & 60.4 & 11.0 & 62.8 & \first{90.2} & 8.5 & 31.1 \\
        \midrule
        Seedance 1.5$^*$ & 31.7 & 8.5 & 65.9 & 7.3 & 47.0 & 54.9 & 11.0 & 20.4 \\
        Kling 3.0$^*$ & 51.8 & 23.2 & 71.3 & 14.6 & 70.1 & 80.5 & 15.9 & \third{33.5} \\
        Seedance 2.0$^*$ & \third{55.5} & \second{24.4} & \first{82.3} & \first{19.5} & \second{71.3} & 64.6 & \first{25.6} & \third{33.5} \\
        Veo 3.1$^*$ & \first{61.6} & \first{29.3} & \third{75.0} & \first{19.5} & \first{82.3} & \second{89.0} & \second{24.4} & \first{40.5} \\
        Grok Imagine$^*$ & \second{59.8} & \second{24.4} & 67.7 & \first{19.5} & \third{70.7} & \third{85.4} & 18.3 & \second{36.9} \\
        \bottomrule
    \end{tabular}
\end{table}

\paragraph{Surface Friction (Table~\ref{tab:prim_surface_friction}).}
Here, the action is the same in both prompts (e.g., hard braking), and only the surface property changes (dry asphalt vs.\ icy road, or smooth glass vs.\ rubber mat). The model needs to understand that friction affects how the same action plays out.
This turns out to be very hard: the best $O_p$ is just 41.4\% (Grok~Imagine), and most models are below 20\%.
The $P_p$ scores are especially revealing---Cosmos-Predict2 scores $P_p = 0.0\%$, meaning none of its video pairs show any physically correct difference in how friction affects the outcome.
The core difficulty is that friction is invisible in the initial frame: a dry road and an icy road can look similar, and models must rely on the text to reason about how surface properties change the physics, rather than copying visual patterns they have seen in training.

\begin{table}[tbp]
    \centering
    \caption{\textbf{Surface Friction} — per-model results (\%). Columns as in Table~\ref{tab:main_results}.}
    \label{tab:prim_surface_friction}
    \renewcommand{\arraystretch}{1.15}
    \begin{tabular}{@{} l cccccccc @{}}
        \toprule
        \textbf{Model} & $A_s$ & $A_p$ & $P_s$ & $P_p$ & $E_s$ & $E_p$ & $O_p$ & \textbf{APEO} \\
        \midrule
        CogVideoX1.5 & 21.7 & 16.7 & 13.3 & 3.3 & 15.0 & 26.7 & 3.3 & 12.5 \\
        Wan2.2 & 31.7 & 23.3 & 43.3 & 6.7 & 50.0 & 36.7 & 6.7 & 18.3 \\
        HunyuanVideo1.5 & 51.7 & 46.7 & 55.0 & 6.7 & 33.3 & 40.0 & 10.0 & 25.8 \\
        Cosmos-Predict2 & 41.7 & 26.7 & 16.7 & 0.0 & 20.0 & 40.0 & 0.0 & 16.7 \\
        \midrule
        Seedance 1.5$^*$ & \third{66.7} & \third{50.0} & \first{63.3} & \third{13.3} & 50.0 & 40.0 & 13.3 & 29.2 \\
        Kling 3.0$^*$ & 60.0 & 46.7 & 43.3 & \third{13.3} & \second{65.0} & \first{70.0} & \second{20.0} & \second{37.5} \\
        Seedance 2.0$^*$ & \second{68.3} & 46.7 & \third{60.0} & \third{13.3} & \first{71.7} & \second{63.3} & \second{20.0} & \third{35.8} \\
        Veo 3.1$^*$ & \third{66.7} & \second{53.3} & 45.0 & \second{16.7} & 48.3 & 53.3 & 13.3 & 34.2 \\
        Grok Imagine$^*$ & \first{74.1} & \first{62.1} & \second{60.3} & \first{34.5} & \third{58.6} & \third{58.6} & \first{41.4} & \first{49.1} \\
        \bottomrule
    \end{tabular}
\end{table}

\paragraph{Material / Medium (Table~\ref{tab:prim_materialmedium}).}
This primitive changes the material properties of the objects or the medium the agent operates in---for example, clear weather vs.\ rain for driving, or a rigid wooden block vs.\ a soft sponge for manipulation.
Veo~3.1 stands out here with 64.2\% APEO and $O_p = 66.7\%$, well ahead of Grok~Imagine (51.7\%).
One possible explanation is that material differences are at least partially visible (wood looks different from sponge), giving models a visual cue to work with---unlike friction, where the causal variable is entirely latent.
The $A_p$ scores are also relatively high for the top models ($\geq$48\%), indicating that models can at least distinguish the actions when interacting with different materials, even if the downstream physics is not always correct.

\begin{table}[tbp]
    \centering
    \caption{\textbf{Material / Medium} — per-model results (\%). Columns as in Table~\ref{tab:main_results}.}
    \label{tab:prim_materialmedium}
    \renewcommand{\arraystretch}{1.15}
    \begin{tabular}{@{} l cccccccc @{}}
        \toprule
        \textbf{Model} & $A_s$ & $A_p$ & $P_s$ & $P_p$ & $E_s$ & $E_p$ & $O_p$ & \textbf{APEO} \\
        \midrule
        CogVideoX1.5 & 31.7 & 26.7 & 25.0 & 0.0 & 40.0 & 53.3 & 0.0 & 20.0 \\
        Wan2.2 & 41.7 & 36.7 & 36.7 & 6.7 & 38.3 & 66.7 & 6.7 & 29.2 \\
        HunyuanVideo1.5 & 45.0 & 30.0 & 45.0 & 13.3 & 36.7 & 70.0 & 20.0 & 33.3 \\
        Cosmos-Predict2 & 60.0 & 36.7 & 41.7 & 13.3 & 60.0 & 63.3 & 13.3 & 31.7 \\
        \midrule
        Seedance 1.5$^*$ & \third{63.3} & \second{50.0} & \third{56.7} & \third{23.3} & 63.3 & \first{83.3} & \third{30.0} & \third{46.7} \\
        Kling 3.0$^*$ & 58.3 & \second{50.0} & 53.3 & 13.3 & \third{75.0} & \second{73.3} & 13.3 & 37.5 \\
        Seedance 2.0$^*$ & 51.7 & \second{50.0} & 53.3 & 16.7 & \first{80.0} & \second{73.3} & 16.7 & 39.2 \\
        Veo 3.1$^*$ & \second{73.3} & \first{70.0} & \first{76.7} & \first{60.0} & 71.7 & 60.0 & \first{66.7} & \first{64.2} \\
        Grok Imagine$^*$ & \first{75.9} & 48.3 & \second{70.7} & \second{37.9} & \second{75.9} & 72.4 & \second{48.3} & \second{51.7} \\
        \bottomrule
    \end{tabular}
\end{table}

\paragraph{Obstacle Configuration (Table~\ref{tab:prim_obstacle_configuration}).}
This primitive tests whether models react differently when the scene layout changes---a clear road vs.\ one with a traffic cone, or a clear workspace vs.\ one with an obstructing object.
Grok~Imagine dominates at 52.5\% APEO, 17\,pp ahead of the runner-up (Veo~3.1, 35.8\%).
A notable pattern in this table is the low $E_p$ scores: even Grok~Imagine only reaches $E_p = 40.0\%$, because changing the obstacle configuration inherently changes what the scene looks like, making it harder for models to keep the non-obstacle parts of the background identical across the pair.
At the bottom of the table, HunyuanVideo and CogVideoX both score $P_p = 0.0\%$ and $O_p \leq 3.3\%$, meaning they essentially cannot handle obstacle-related reasoning at all.

\begin{table}[tbp]
    \centering
    \caption{\textbf{Obstacle Configuration} — per-model results (\%). Columns as in Table~\ref{tab:main_results}.}
    \label{tab:prim_obstacle_configuration}
    \renewcommand{\arraystretch}{1.15}
    \begin{tabular}{@{} l cccccccc @{}}
        \toprule
        \textbf{Model} & $A_s$ & $A_p$ & $P_s$ & $P_p$ & $E_s$ & $E_p$ & $O_p$ & \textbf{APEO} \\
        \midrule
        CogVideoX1.5 & 35.0 & 16.7 & 25.0 & 0.0 & 13.3 & 23.3 & 3.3 & 10.8 \\
        Wan2.2 & 43.3 & 23.3 & 41.7 & 10.0 & 45.0 & 33.3 & 3.3 & 17.5 \\
        HunyuanVideo1.5 & 48.3 & 30.0 & 45.0 & 0.0 & 33.3 & 36.7 & 0.0 & 16.7 \\
        Cosmos-Predict2 & 33.3 & 23.3 & 31.7 & 6.7 & 58.3 & \first{60.0} & 3.3 & 23.3 \\
        \midrule
        Seedance 1.5$^*$ & 60.0 & \third{36.7} & 40.0 & 13.3 & 45.0 & 40.0 & \third{20.0} & 27.5 \\
        Kling 3.0$^*$ & \third{65.0} & \second{53.3} & 45.0 & \third{20.0} & 50.0 & \second{53.3} & 13.3 & \third{35.0} \\
        Seedance 2.0$^*$ & \third{65.0} & \third{36.7} & \third{51.7} & 6.7 & \first{70.0} & \third{43.3} & 6.7 & 23.3 \\
        Veo 3.1$^*$ & \second{68.3} & \third{36.7} & \second{56.7} & \second{26.7} & \second{65.0} & \third{43.3} & \second{36.7} & \second{35.8} \\
        Grok Imagine$^*$ & \first{86.7} & \first{66.7} & \first{63.3} & \first{50.0} & \third{61.7} & 40.0 & \first{53.3} & \first{52.5} \\
        \bottomrule
    \end{tabular}
\end{table}

\subsection{Per-Dimension Failure Analysis}
\label{sec:failure_analysis}

We analyze the VLM judge's reasoning across all 2{,}871 evaluated instances to characterize how each APEO dimension fails under single-video and paired evaluation.
Table~\ref{tab:failure_modes} summarizes the dominant failure modes and their prevalence.

\begin{table}[t]
    \centering
    \caption{\textbf{Failure mode taxonomy.} For each APEO dimension, we classify failures into modes based on VLM judge reasoning. Percentages indicate the fraction of failures in that dimension exhibiting each mode (categories can overlap). $N$ = number of failing instances.}
    \label{tab:failure_modes}

    \renewcommand{\arraystretch}{1.2}
    \begin{tabular}{@{} l l l r @{}}
        \toprule
        \textbf{Dim.} & \textbf{Eval} & \textbf{Failure Mode} & \textbf{\%} \\
        \midrule
        \multirow{5}{*}{$A$}
        & \multirow{2}{*}{Single ($N{=}2007$)}
        & Action omission (ignores instruction) & 71.1 \\
        & & Action substitution (wrong action) & 34.3 \\
        \cmidrule(l){2-4}
        & \multirow{2}{*}{Paired ($N{=}1940$)}
        & No discernible inter-video difference & 63.9 \\
        & & Baseline inconsistency (state diverges) & 39.4 \\
        \midrule
        \multirow{6}{*}{$P$}
        & \multirow{3}{*}{Single ($N{=}1568$)}
        & Kinematic discontinuity / teleportation & 32.0 \\
        & & Object / arm morphing & 28.8 \\
        & & Ghost force / levitation & 26.6 \\
        \cmidrule(l){2-4}
        & \multirow{2}{*}{Paired ($N{=}2194$)}
        & Mode collapse (identical trajectories) & 90.0 \\
        & & No interaction occurs & 5.4 \\
        \midrule
        \multirow{5}{*}{$E$}
        & \multirow{3}{*}{Single ($N{=}1691$)}
        & Object permanence violation & 70.4 \\
        & & Background collapse / distortion & 33.9 \\
        & & Camera hallucination & 18.8 \\
        \cmidrule(l){2-4}
        & \multirow{2}{*}{Paired ($N{=}1020$)}
        & Cross-video environment divergence & 73.9 \\
        & & Divergent hallucinations & 26.4 \\
        \midrule
        \multirow{3}{*}{$O$}
        & \multirow{3}{*}{Paired ($N{=}2159$)}
        & Mode collapse (same final state) & 59.4 \\
        & & Upstream propagation (action never happened) & 50.3 \\
        & & Causal inversion (wrong direction) & 9.8 \\
        \bottomrule
    \end{tabular}
\end{table}

\paragraph{Adherence ($A$).}
$A_s$ fails when a video does not execute the prompted action at all.
The dominant mode is \emph{action omission}: the model generates a plausible driving or manipulation clip while ignoring the instruction---the judge observes \textit{``the ego car maintains constant speed''} when acceleration was prompted, or \textit{``the gripper lowers directly over the target''} when a spatial offset was specified.
A secondary mode is \emph{action substitution}: the model executes a different action (e.g., the front car accelerates away instead of the ego car accelerating forward).
$A_p$ fails differently---it requires that the \emph{difference} between the two videos' actions be visually unambiguous.
The most common $A_p$ failure is \emph{baseline inconsistency}: both videos individually show plausible actions, but their arm kinematics or vehicle trajectories differ from the very first frame, making it impossible to attribute any divergence to the causal variable rather than to uncontrolled generation variance.
The judge notes: \textit{``the arm approach trajectories are completely different between the two videos---a viewer could not identify which spatial difference is due to the prompt.''}

\paragraph{Physics ($P$).}
$P_s$ fails when a single video contains internal physical violations---object morphing, teleportation, or impossible kinematics.
Typical observations include: \textit{``the robotic arm exhibits severe morphing and loses structural integrity,''} \textit{``the lid distorts into an unrecognizable blob,''} and \textit{``a duplicate object spontaneously spawns mid-video.''}
These are generation-quality failures unrelated to causal reasoning.
$P_p$ failures are fundamentally different: each video may look physically plausible on its own, but their trajectories do not bifurcate at the expected intervention point.
The judge identifies three recurring patterns: (i)~\emph{mode collapse}---\textit{``the gap closes in both videos at the same rate; there is no trajectory divergence''}; (ii)~\emph{causal inversion}---\textit{``the gap in Video~B is larger than in Video~A, representing a reversed physical relationship''}; and (iii)~\emph{frozen environment}---\textit{``the surface friction variable produced no causal effect on the object's displacement.''}
Mode collapse accounts for the majority of $P_p$ failures and is the core mechanism behind the $P_s$--$P_p$ gap: models produce individually plausible physics that is simply copied across both conditions.

\paragraph{Environment ($E$).}
$E_s$ fails when the background or non-target objects within a single video are unstable.
The judge reports \emph{background collapse} (\textit{``the building on the right liquefies and distorts''}), \emph{object permanence violations} (\textit{``the parked excavator morphs into a moving truck''}), and \emph{unscripted object spawns} (\textit{``a duplicate donut appears on the table while the arm holds another''}).
$E_p$ requires that the two videos share an identical ``canvas''---same background, same non-target objects, same camera.
It fails when generation variance introduces \emph{divergent hallucinations}: \textit{``Video~B introduces a green barrier and changes the curb style in areas that should be identical to Video~A,''} or \textit{``the workspace background and camera angle are completely inconsistent between the two videos.''}

\paragraph{Outcome ($O$).}
$O_p$ measures whether the expected causal state divergence is observable between the pair.
Failures fall into three categories.
\emph{Mode collapse} (most common): both videos reach the same final state---\textit{``both videos show the cap remaining on the bed; neither grasps it''} or \textit{``the front car appears the same size at the end of both videos; no gap difference emerges.''}
\emph{Upstream propagation}: an earlier failure (action omission, object morphing) prevents any meaningful outcome---\textit{``because the arm fails to make contact in either video, the plate does not move in either case.''}
\emph{Causal inversion}: the outcome diverges in the wrong direction---\textit{``the acceleration instruction resulted in a widening gap''} or \textit{``the stopping distances are reversed between conditions.''}
The relative prevalence shifts by primitive: $\mathcal{D}_\text{int}$ failures are dominated by upstream propagation (the action was never executed), while $\mathcal{D}_\text{env}$ failures are dominated by mode collapse (the action was correct but the environment change had no physical effect).

\textbf{Model rankings shift across primitives, with no single model dominating.}
Grok~Imagine leads on Force/Degree (70.7\% APEO), Surface Friction (49.1\%), and Obstacle Configuration (52.5\%); Veo~3.1 leads on Spatial Alignment (60.8\%), Temporal Sequencing (40.5\%), and Material/Medium (64.2\%).
Some specializations are extreme: Grok scores 52.5\% on Obstacle Configuration, where the next-best model (Veo) scores 35.8\% (a 17\,pp lead), yet falls below Veo on Temporal Sequencing.
If models exhibited general causal sensitivity, performance would transfer across related tasks within the same primitive.
Instead, Grok scores 0\% on Longitudinal Braking but 83\% on Longitudinal Acceleration, both within the \emph{same} Spatial Alignment primitive.
These spiky profiles indicate that task-specific success does not generalize across compositionally related tasks within a primitive, a pattern more consistent with training-distribution familiarity than with systematic physical sensitivity.

\subsection{Per-Sub-Category Outcome Scores}
\label{app:heatmap}

Table~\ref{tab:heatmap} provides the most granular view of model performance: every model's outcome score ($O_p$) on every sub-category, sorted from easiest (left) to hardest (right).
The color coding makes several patterns immediately visible.

\textbf{No model is uniformly strong.}
Even the top-ranked Grok~Imagine has two sub-categories at 0\% (Longitudinal Braking and Overtake with Acceleration)---tasks on which it completely fails despite achieving 100\% on Acceleration Intensity.
This extreme spikiness (from 0\% to 100\% within the same model) rules out the possibility that high aggregate scores reflect general causal reasoning; instead, they reflect a patchwork of task-specific capabilities.

\textbf{The right side of the table is almost entirely blue or white.}
Sub-categories like Object Mass (hardest, 7.4\% average), Overtake with Acceleration (9.6\%), and Table Friction (13.3\%) are near-zero for most models.
These tasks share a common property: the causal variable is either invisible (mass), latent (friction coefficient), or requires relational reasoning across agents (timing).
In contrast, the left side of the table (Acceleration Intensity, Longitudinal Acceleration) involves visually dramatic, training-familiar scenarios where models achieve 50--100\%.

\textbf{Open-source models (bottom rows) show a clear ceiling.}
Their scores rarely exceed the \colorbox{rank2}{orange} tier (25--49\%), and the vast majority of cells on the right half are white (0\%).
Closed-source models populate more \colorbox{rank1}{red} and \colorbox{rank2}{orange} cells, but even they collapse on the hardest sub-categories.

\begin{table*}[htbp]

    \centering

    \caption{\textbf{Per-model $O_p$ (\%) on every sub-category,} sorted by difficulty (easiest top). Cells are color-coded: \colorbox{rank1}{$\geq$50}, \colorbox{rank2}{25--49}, \colorbox{rank3}{1--24}, white = 0.}

    \label{tab:heatmap}

    \tiny

    \renewcommand{\arraystretch}{1.2}

    \begin{tabular}{@{} l ccccccccc @{}}

        \toprule

        \textbf{Sub-category} & \textbf{Grok} & \textbf{Veo 3.1} & \textbf{Seedance 2.0} & \textbf{Seedance 1.5} & \textbf{Kling} & \textbf{Wan} & \textbf{Cosmos-Predict2} & \textbf{CogVideoX} & \textbf{Hunyuan} \\

        \midrule

        Accel Intensity & \cellcolor{rank1}100 & \cellcolor{rank1}93 & \cellcolor{rank1}100 & \cellcolor{rank1}93 & \cellcolor{rank1}79 & \cellcolor{rank1}64 & \cellcolor{rank1}71 & \cellcolor{rank2}36 & \cellcolor{rank2}43 \\

        Long. Accel & \cellcolor{rank1}83 & \cellcolor{rank1}100 & \cellcolor{rank1}50 & \cellcolor{rank1}56 & \cellcolor{rank2}28 & \cellcolor{rank1}61 & \cellcolor{rank3}17 & \cellcolor{rank1}56 & \cellcolor{rank3}6 \\

        Brake Intensity & \cellcolor{rank1}78 & \cellcolor{rank1}67 & \cellcolor{rank1}50 & \cellcolor{rank2}39 & \cellcolor{rank3}11 & \cellcolor{rank3}6 & \cellcolor{rank3}17 & \cellcolor{rank3}11 & \cellcolor{rank3}6 \\

        Release Coords & \cellcolor{rank1}70 & \cellcolor{rank1}60 & \cellcolor{rank2}35 & \cellcolor{rank2}35 & \cellcolor{rank3}15 & \cellcolor{rank3}10 & \cellcolor{rank3}15 & \cellcolor{rank3}10 & \cellcolor{rank3}20 \\

        Deformability & \cellcolor{rank1}53 & \cellcolor{rank1}60 & \cellcolor{rank2}27 & \cellcolor{rank2}40 & \cellcolor{rank2}27 & \cellcolor{rank3}7 & \cellcolor{rank2}27 & 0 & \cellcolor{rank2}27 \\

        Grasp Align & \cellcolor{rank2}36 & \cellcolor{rank1}55 & \cellcolor{rank2}41 & \cellcolor{rank3}14 & \cellcolor{rank2}41 & \cellcolor{rank3}14 & \cellcolor{rank2}36 & \cellcolor{rank3}5 & \cellcolor{rank3}14 \\

        Long. Brake & 0 & \cellcolor{rank2}28 & \cellcolor{rank1}50 & \cellcolor{rank2}39 & \cellcolor{rank3}22 & \cellcolor{rank2}33 & \cellcolor{rank3}17 & \cellcolor{rank3}17 & \cellcolor{rank2}28 \\

        Push/Yank & \cellcolor{rank1}55 & \cellcolor{rank2}30 & \cellcolor{rank2}25 & \cellcolor{rank2}40 & \cellcolor{rank3}10 & \cellcolor{rank3}10 & \cellcolor{rank2}25 & \cellcolor{rank3}5 & \cellcolor{rank3}15 \\

        Scene Layout & \cellcolor{rank1}80 & \cellcolor{rank2}47 & \cellcolor{rank3}13 & \cellcolor{rank2}33 & \cellcolor{rank2}27 & \cellcolor{rank3}7 & 0 & \cellcolor{rank3}7 & 0 \\

        Gripper Timing & \cellcolor{rank3}18 & \cellcolor{rank3}23 & \cellcolor{rank2}36 & \cellcolor{rank3}14 & \cellcolor{rank2}32 & \cellcolor{rank2}36 & \cellcolor{rank3}23 & \cellcolor{rank3}9 & \cellcolor{rank3}5 \\

        Lateral Avoid & \cellcolor{rank1}59 & \cellcolor{rank1}59 & \cellcolor{rank2}35 & 0 & \cellcolor{rank3}24 & 0 & 0 & 0 & 0 \\

        Overtake Brake & \cellcolor{rank3}20 & \cellcolor{rank3}7 & \cellcolor{rank2}40 & \cellcolor{rank3}20 & \cellcolor{rank2}27 & \cellcolor{rank3}20 & \cellcolor{rank3}7 & \cellcolor{rank3}20 & \cellcolor{rank3}7 \\

        Road Medium & \cellcolor{rank2}43 & \cellcolor{rank1}73 & \cellcolor{rank3}7 & \cellcolor{rank3}20 & 0 & \cellcolor{rank3}7 & 0 & 0 & \cellcolor{rank3}13 \\

        Road Friction & \cellcolor{rank2}43 & \cellcolor{rank3}7 & \cellcolor{rank3}13 & \cellcolor{rank3}20 & \cellcolor{rank3}20 & \cellcolor{rank3}7 & 0 & \cellcolor{rank3}7 & \cellcolor{rank3}20 \\

        Table Friction & \cellcolor{rank2}40 & \cellcolor{rank3}20 & \cellcolor{rank2}27 & \cellcolor{rank3}7 & \cellcolor{rank3}20 & \cellcolor{rank3}7 & 0 & 0 & 0 \\

        Overtake Hard & \cellcolor{rank2}27 & \cellcolor{rank2}47 & \cellcolor{rank3}3 & \cellcolor{rank3}10 & \cellcolor{rank3}3 & \cellcolor{rank3}10 & \cellcolor{rank3}3 & \cellcolor{rank3}3 & 0 \\

        Overtake Accel & 0 & 0 & \cellcolor{rank2}40 & 0 & \cellcolor{rank3}7 & \cellcolor{rank3}13 & 0 & \cellcolor{rank2}27 & 0 \\

        Object Mass & \cellcolor{rank2}27 & \cellcolor{rank2}27 & 0 & \cellcolor{rank3}7 & 0 & 0 & \cellcolor{rank3}7 & 0 & 0 \\

        \bottomrule

    \end{tabular}

\end{table*}

\subsection{AD vs.\ Robotics Domain Comparison}
\label{app:domain}

A key question for any embodied benchmark is whether the findings are domain-specific or generalizable.
Table~\ref{tab:domain_comparison} splits each model's paired scores by domain: Autonomous Driving (AD) and Robotic Manipulation (Rob).

Most models perform better on AD than Robotics, which is expected---driving videos are far more common in web-scale training data.
The gap is largest for CogVideoX1.5 (+17.7\,pp) and Seedance~2.0 (+12.7\,pp), suggesting these models' apparent capabilities are heavily biased toward road scenarios.
However, Cosmos-Predict2 shows the \emph{opposite} pattern: it scores 11.1\,pp higher on Robotics than AD, likely reflecting its origin as an embodied world model with manipulation-relevant training data.

The two top models---Grok~Imagine and Veo~3.1---are remarkably domain-balanced ($\Delta < 2.2$\,pp).
Their causal reasoning capability, while imperfect, transfers across embodiments rather than being tied to one visual domain.
This suggests that the best current models have at least partially learned domain-general physical reasoning, even though they still fail on most tasks.

Interestingly, $E_p$ (environment consistency) is consistently higher in AD than in Robotics for most models, likely because driving scenes have simpler, more predictable backgrounds (roads, lane markings) compared to cluttered tabletop manipulation environments.

\begin{table*}[htbp]
    \centering
    \caption{\textbf{Per-model APEO scores (\%) split by domain.} AD = Autonomous Driving (11 sub-categories); Rob = Robotic Manipulation (7 sub-categories). $\Delta$ = AD $-$ Rob. $^*$ = closed-source.}
    \label{tab:domain_comparison}
    \small
    \renewcommand{\arraystretch}{1.2}
    \begin{tabular}{@{} l cccc c cccc c c @{}}
        \toprule
        & \multicolumn{4}{c}{\textbf{AD}} && \multicolumn{4}{c}{\textbf{Robotics}} && \\
        \cmidrule(lr){2-5}\cmidrule(lr){7-10}
        \textbf{Model} & $A_p$ & $P_p$ & $E_p$ & APEO && $A_p$ & $P_p$ & $E_p$ & APEO && $\Delta$APEO \\
        \midrule
        CogVideoX1.5 & 23.7 & 13.7 & 66.3 & 29.9 && 10.1 & 6.2 & 27.9 & 12.2 && +17.7 \\
        Wan2.2 & 25.3 & 14.7 & 68.4 & 32.1 && 17.8 & 13.2 & 48.1 & 23.1 && +9.0 \\
        HunyuanVideo1.5 & 19.5 & 9.5 & 72.1 & 27.8 && 21.7 & 16.3 & 54.3 & 26.0 && +1.8 \\
        Cosmos-Predict2 & 17.4 & 10.5 & 74.7 & 28.4 && 36.4 & 31.0 & 70.5 & 39.5 && $-$11.1 \\
        \midrule
        Seedance 1.5$^*$ & 41.1 & 22.6 & 57.9 & 37.5 && 18.6 & 17.8 & 41.1 & 25.0 && +12.5 \\
        Kling 3.0$^*$ & 38.4 & 20.0 & 84.7 & 40.9 && 29.5 & 29.5 & 60.5 & 35.3 && +5.6 \\
        Seedance 2.0$^*$ & 49.5 & 32.1 & 72.1 & 47.0 && 30.2 & 24.0 & 54.3 & 34.3 && +12.7 \\
        Veo 3.1$^*$ & 51.1 & 38.9 & 66.8 & 51.3 && 43.4 & 41.9 & 76.0 & 50.2 && +1.1 \\
        Grok Imagine$^*$ & 50.0 & 41.5 & 71.8 & 52.5 && 48.1 & 44.2 & 66.7 & 50.4 && +2.1 \\
        \bottomrule
    \end{tabular}
\end{table*}

\subsection{Closed-Source vs.\ Open-Source Gap by Primitive}
\label{app:cs_os}

Table~\ref{tab:cs_os_gap} breaks down the closed-source (CS) vs.\ open-source (OS) performance gap by physical primitive.
The overall gap is 15.3\,pp, but this number hides substantial variation across primitives.

\textbf{Force/Degree shows the largest gap (+23.2\,pp).}
This primitive tests whether models can produce different magnitudes of the same action (e.g., gentle vs.\ hard braking).
Commercial models likely benefit from both larger model capacity and more diverse training data covering a wide range of force-outcome scenarios.

\textbf{Temporal Sequencing shows the smallest gap (+5.9\,pp).}
This is arguably the most important finding in this table: the primitive that requires coordinating timing across multiple agents is equally difficult for all models regardless of scale or data.
The near-zero gap suggests that temporal reasoning is not a problem that more training data or larger models can solve in the current paradigm---it may require architectural innovations (e.g., explicit relational reasoning modules, or physical simulation layers).

The remaining four primitives fall in a narrow band (16--19\,pp gap), suggesting that for spatial, environmental, and material reasoning, commercial scale provides a consistent but moderate advantage.

\begin{table}[htbp]
    \centering
    \caption{\textbf{APEO (\%) by model tier and primitive.} CS = closed-source (5 models); OS = open-source (4 models). The gap is largest for Force/Degree and smallest for Temporal Sequencing.}
    \label{tab:cs_os_gap}
    \small
    \renewcommand{\arraystretch}{1.2}
    \begin{tabular}{@{} l ccc @{}}
        \toprule
        \textbf{Primitive} & \textbf{CS} & \textbf{OS} & \textbf{Gap} \\
        \midrule
        Spatial Alignment & 48.1 & 32.0 & +16.0 \\
        Force / Degree & 55.4 & 32.2 & +23.2 \\
        Temporal Sequencing & 33.0 & 27.1 & +5.9 \\
        Surface Friction & 37.1 & 18.3 & +18.8 \\
        Material / Medium & 47.8 & 28.5 & +19.3 \\
        Obstacle Configuration & 34.8 & 17.1 & +17.7 \\
        \midrule
        \textbf{Overall} & \textbf{43.1} & \textbf{27.8} & \textbf{+15.3} \\
        \bottomrule
    \end{tabular}
\end{table}

\subsection{Detailed Per-Primitive Results}
\label{app:detailed_primitive}
\begin{itemize}[nosep,leftmargin=*]
    \item \textbf{Extreme task-level variance within a single model.} Grok scores $A_p = 83.3\%$ on Longitudinal Acceleration but $A_p = 0.0\%$ on Longitudinal Braking---within the same Spatial Alignment primitive. Similarly, Veo~3.1 achieves $O_p = 100\%$ on Longitudinal Acceleration but $O_p = 0\%$ on Overtake with Acceleration.
    \item \textbf{$P_s$ can be near-perfect while $P_p$ is zero.} HunyuanVideo achieves $P_s = 100\%$ on Overtake Hard (every individual video has plausible physics) but $P_p = 0\%$ (the two videos in each pair are physically interchangeable). This is the ``individually plausible, causally meaningless'' failure mode at its most extreme.
\end{itemize}

\begin{table*}[htbp]
    \centering
    \caption{\textbf{Spatial Alignment (Autonomous Driving) — per-sub-category results (\%).} Full APEO breakdown for each model on every AD sub-category within this primitive. \textbf{Bold}: best per column within each sub-category.}
    \label{tab:app_detail_spatial_alignment_ad}
    \small
    \renewcommand{\arraystretch}{1.15}
    \begin{tabular}{@{} l cccccccc @{}}
        \toprule
        \textbf{Model} & $A_s$ & $A_p$ & $P_s$ & $P_p$ & $E_s$ & $E_p$ & $O_p$ & \textbf{APEO} \\
        \midrule
        \multicolumn{9}{l}{\textit{Long. Accel (AD)}} \\
        \midrule
        CogVideoX1.5 & 72.2 & 55.6 & 80.6 & 55.6 & 41.7 & 44.4 & 55.6 & 52.8 \\
        Wan2.2 & 69.4 & 50.0 & 69.4 & 33.3 & 58.3 & \textbf{83.3} & 61.1 & 56.9 \\
        HunyuanVideo1.5 & 47.2 & 5.6 & 91.7 & 5.6 & 52.8 & 66.7 & 5.6 & 20.8 \\
        Cosmos-Predict2 & 38.9 & 16.7 & 94.4 & 16.7 & 52.8 & \textbf{83.3} & 16.7 & 33.3 \\
        \midrule
        Seedance 1.5$^*$ & 69.4 & 55.6 & 69.4 & 33.3 & 58.3 & 33.3 & 55.6 & 44.4 \\
        Kling 3.0$^*$ & 61.1 & 27.8 & 94.4 & 22.2 & \textbf{80.6} & 77.8 & 27.8 & 38.9 \\
        Seedance 2.0$^*$ & 75.0 & 50.0 & 94.4 & 38.9 & 72.2 & 66.7 & 50.0 & 51.4 \\
        Veo 3.1$^*$ & \textbf{94.4} & \textbf{94.4} & \textbf{97.2} & \textbf{94.4} & 77.8 & 55.6 & \textbf{100.0} & \textbf{86.1} \\
        Grok Imagine$^*$ & 80.6 & 83.3 & 88.9 & 83.3 & 77.8 & 77.8 & 83.3 & 81.9 \\
        \midrule
        \multicolumn{9}{l}{\textit{Long. Brake (AD)}} \\
        \midrule
        CogVideoX1.5 & 38.9 & 11.1 & 66.7 & 11.1 & 41.7 & 61.1 & 16.7 & 25.0 \\
        Wan2.2 & 50.0 & 27.8 & 88.9 & 27.8 & \textbf{77.8} & 88.9 & 33.3 & 44.4 \\
        HunyuanVideo1.5 & 55.6 & 27.8 & \textbf{100.0} & 27.8 & 72.2 & 72.2 & 27.8 & 38.9 \\
        Cosmos-Predict2 & 22.2 & 16.7 & 83.3 & 11.1 & 69.4 & 72.2 & 16.7 & 29.2 \\
        \midrule
        Seedance 1.5$^*$ & 55.6 & \textbf{38.9} & 80.6 & 22.2 & 69.4 & 72.2 & 38.9 & 43.1 \\
        Kling 3.0$^*$ & 58.3 & 22.2 & 91.7 & 16.7 & \textbf{77.8} & \textbf{94.4} & 22.2 & 38.9 \\
        Seedance 2.0$^*$ & \textbf{69.4} & \textbf{38.9} & 97.2 & \textbf{33.3} & 72.2 & 61.1 & \textbf{50.0} & \textbf{45.8} \\
        Veo 3.1$^*$ & 61.1 & 27.8 & 72.2 & 27.8 & 75.0 & 66.7 & 27.8 & 37.5 \\
        Grok Imagine$^*$ & 27.8 & 0.0 & 69.4 & 0.0 & 75.0 & 33.3 & 0.0 & 8.3 \\
        \midrule
        \multicolumn{9}{l}{\textit{Lateral Avoid (AD)}} \\
        \midrule
        CogVideoX1.5 & 11.8 & 11.8 & 14.7 & 11.8 & 17.6 & 64.7 & 0.0 & 22.1 \\
        Wan2.2 & 11.8 & 11.8 & 5.9 & 11.8 & 20.6 & 64.7 & 0.0 & 22.1 \\
        HunyuanVideo1.5 & 35.3 & 17.6 & 17.6 & 17.6 & 35.3 & 70.6 & 0.0 & 26.5 \\
        Cosmos-Predict2 & 17.6 & 23.5 & 17.6 & 17.6 & 55.9 & 76.5 & 0.0 & 29.4 \\
        \midrule
        Seedance 1.5$^*$ & 32.4 & 23.5 & 20.6 & 17.6 & 17.6 & 41.2 & 0.0 & 20.6 \\
        Kling 3.0$^*$ & 58.8 & 52.9 & 47.1 & 52.9 & \textbf{79.4} & 82.4 & 23.5 & 52.9 \\
        Seedance 2.0$^*$ & \textbf{64.7} & \textbf{70.6} & 50.0 & \textbf{70.6} & 58.8 & \textbf{94.1} & 35.3 & \textbf{67.6} \\
        Veo 3.1$^*$ & 58.8 & 35.3 & 52.9 & 35.3 & 38.2 & 82.4 & \textbf{58.8} & 52.9 \\
        Grok Imagine$^*$ & \textbf{64.7} & 58.8 & \textbf{64.7} & 58.8 & 50.0 & 64.7 & \textbf{58.8} & 60.3 \\
        \bottomrule
    \end{tabular}
\end{table*}

\begin{table*}[htbp]
    \centering
    \caption{\textbf{Spatial Alignment (Robotic Manipulation) — per-sub-category results (\%).} Full APEO breakdown for each model on every Rob sub-category within this primitive. \textbf{Bold}: best per column within each sub-category.}
    \label{tab:app_detail_spatial_alignment_rob}
    \small
    \renewcommand{\arraystretch}{1.15}
    \begin{tabular}{@{} l cccccccc @{}}
        \toprule
        \textbf{Model} & $A_s$ & $A_p$ & $P_s$ & $P_p$ & $E_s$ & $E_p$ & $O_p$ & \textbf{APEO} \\
        \midrule
        \multicolumn{9}{l}{\textit{Release Coords (Rob)}} \\
        \midrule
        CogVideoX1.5 & 17.5 & 10.0 & 25.0 & 20.0 & 20.0 & 55.0 & 10.0 & 23.8 \\
        Wan2.2 & 32.5 & 30.0 & 42.5 & 20.0 & 42.5 & 60.0 & 10.0 & 30.0 \\
        HunyuanVideo1.5 & 40.0 & 20.0 & 75.0 & 40.0 & 50.0 & 70.0 & 20.0 & 37.5 \\
        Cosmos-Predict2 & 40.0 & 30.0 & 77.5 & 60.0 & 65.0 & \textbf{90.0} & 15.0 & 48.8 \\
        \midrule
        Seedance 1.5$^*$ & 67.5 & 45.0 & 72.5 & 40.0 & 47.5 & 50.0 & 35.0 & 42.5 \\
        Kling 3.0$^*$ & 27.5 & 20.0 & 65.0 & 50.0 & 52.5 & 85.0 & 15.0 & 42.5 \\
        Seedance 2.0$^*$ & 70.0 & 35.0 & 82.5 & 35.0 & 70.0 & 50.0 & 35.0 & 38.8 \\
        Veo 3.1$^*$ & 65.0 & 65.0 & 82.5 & 65.0 & \textbf{90.0} & \textbf{90.0} & 60.0 & 70.0 \\
        Grok Imagine$^*$ & \textbf{87.5} & \textbf{75.0} & \textbf{87.5} & \textbf{80.0} & 85.0 & 85.0 & \textbf{70.0} & \textbf{77.5} \\
        \midrule
        \multicolumn{9}{l}{\textit{Grasp Align (Rob)}} \\
        \midrule
        CogVideoX1.5 & 31.8 & 22.7 & 6.8 & 13.6 & 4.5 & 4.5 & 4.5 & 11.4 \\
        Wan2.2 & 27.3 & 9.1 & 40.9 & 13.6 & 40.9 & 40.9 & 13.6 & 19.3 \\
        HunyuanVideo1.5 & 40.9 & 13.6 & 63.6 & 36.4 & 47.7 & 36.4 & 13.6 & 25.0 \\
        Cosmos-Predict2 & \textbf{68.2} & \textbf{45.5} & 63.6 & 50.0 & 54.5 & 54.5 & 36.4 & 46.6 \\
        \midrule
        Seedance 1.5$^*$ & 40.9 & 9.1 & 45.5 & 13.6 & 31.8 & 18.2 & 13.6 & 13.6 \\
        Kling 3.0$^*$ & 65.9 & \textbf{45.5} & 65.9 & 50.0 & 50.0 & 54.5 & 40.9 & 47.7 \\
        Seedance 2.0$^*$ & 54.5 & 36.4 & 75.0 & 40.9 & 72.7 & 50.0 & 40.9 & 42.0 \\
        Veo 3.1$^*$ & \textbf{68.2} & \textbf{45.5} & \textbf{79.5} & \textbf{63.6} & \textbf{77.3} & \textbf{63.6} & \textbf{54.5} & \textbf{56.8} \\
        Grok Imagine$^*$ & 56.8 & 36.4 & 70.5 & 40.9 & 70.5 & \textbf{63.6} & 36.4 & 44.3 \\
        \bottomrule
    \end{tabular}
\end{table*}

\begin{table*}[htbp]
    \centering
    \caption{\textbf{Force / Degree — per-sub-category results (\%).} Full APEO breakdown for each model on every sub-category within this primitive. \textbf{Bold}: best per column within each sub-category.}
    \label{tab:app_detail_forcedegree}
    \small
    \renewcommand{\arraystretch}{1.15}
    \begin{tabular}{@{} l cccccccc @{}}
        \toprule
        \textbf{Model} & $A_s$ & $A_p$ & $P_s$ & $P_p$ & $E_s$ & $E_p$ & $O_p$ & \textbf{APEO} \\
        \midrule
        \multicolumn{9}{l}{\textit{Brake Intensity (AD)}} \\
        \midrule
        CogVideoX1.5 & 50.0 & 11.1 & \textbf{94.4} & 11.1 & 50.0 & 50.0 & 11.1 & 20.8 \\
        Wan2.2 & 25.0 & 0.0 & 50.0 & 0.0 & 38.9 & 27.8 & 5.6 & 8.3 \\
        HunyuanVideo1.5 & 44.4 & 0.0 & 88.9 & 0.0 & 58.3 & \textbf{83.3} & 5.6 & 22.2 \\
        Cosmos-Predict2 & 55.6 & 11.1 & 83.3 & 11.1 & 61.1 & 55.6 & 16.7 & 23.6 \\
        \midrule
        Seedance 1.5$^*$ & 50.0 & 38.9 & 58.3 & 38.9 & 50.0 & 44.4 & 38.9 & 40.3 \\
        Kling 3.0$^*$ & 44.4 & 11.1 & 83.3 & 11.1 & 72.2 & 72.2 & 11.1 & 26.4 \\
        Seedance 2.0$^*$ & 66.7 & 50.0 & 91.7 & 38.9 & \textbf{80.6} & 77.8 & 50.0 & 54.2 \\
        Veo 3.1$^*$ & 75.0 & \textbf{61.1} & 77.8 & 55.6 & \textbf{80.6} & 55.6 & 66.7 & 59.7 \\
        Grok Imagine$^*$ & \textbf{77.8} & \textbf{61.1} & 88.9 & \textbf{66.7} & 75.0 & 55.6 & \textbf{77.8} & \textbf{65.3} \\
        \midrule
        \multicolumn{9}{l}{\textit{Accel Intensity (AD)}} \\
        \midrule
        CogVideoX1.5 & 53.6 & 35.7 & 82.1 & 35.7 & 32.1 & 92.9 & 35.7 & 50.0 \\
        Wan2.2 & 64.3 & 57.1 & 78.6 & 57.1 & 57.1 & \textbf{100.0} & 64.3 & 69.6 \\
        HunyuanVideo1.5 & 60.7 & 42.9 & 71.4 & 28.6 & 50.0 & 85.7 & 42.9 & 50.0 \\
        Cosmos-Predict2 & 75.0 & 64.3 & 67.9 & 57.1 & 50.0 & 85.7 & 71.4 & 69.6 \\
        \midrule
        Seedance 1.5$^*$ & 85.7 & 85.7 & 82.1 & 71.4 & 57.1 & 85.7 & 92.9 & 83.9 \\
        Kling 3.0$^*$ & 82.1 & 78.6 & 85.7 & 64.3 & 75.0 & \textbf{100.0} & 78.6 & 80.4 \\
        Seedance 2.0$^*$ & \textbf{100.0} & \textbf{100.0} & \textbf{96.4} & \textbf{92.9} & \textbf{78.6} & 92.9 & \textbf{100.0} & \textbf{96.4} \\
        Veo 3.1$^*$ & 82.1 & 92.9 & 78.6 & 85.7 & 46.4 & \textbf{100.0} & 92.9 & 92.9 \\
        Grok Imagine$^*$ & 92.9 & \textbf{100.0} & 89.3 & 85.7 & 42.9 & \textbf{100.0} & \textbf{100.0} & \textbf{96.4} \\
        \midrule
        \multicolumn{9}{l}{\textit{Push/Yank (Rob)}} \\
        \midrule
        CogVideoX1.5 & 20.0 & 15.0 & 5.0 & 0.0 & 5.0 & 35.0 & 5.0 & 13.8 \\
        Wan2.2 & 30.0 & 10.0 & 22.5 & 5.0 & 25.0 & 55.0 & 10.0 & 20.0 \\
        HunyuanVideo1.5 & 35.0 & 20.0 & 37.5 & 10.0 & 32.5 & 65.0 & 15.0 & 27.5 \\
        Cosmos-Predict2 & 37.5 & 25.0 & 52.5 & 20.0 & 52.5 & \textbf{85.0} & 25.0 & 38.8 \\
        \midrule
        Seedance 1.5$^*$ & 50.0 & 15.0 & 40.0 & 20.0 & 35.0 & 45.0 & 40.0 & 30.0 \\
        Kling 3.0$^*$ & 20.0 & 10.0 & 60.0 & 10.0 & 52.5 & 65.0 & 10.0 & 23.8 \\
        Seedance 2.0$^*$ & 55.0 & 35.0 & \textbf{65.0} & 20.0 & \textbf{57.5} & 65.0 & 25.0 & 36.2 \\
        Veo 3.1$^*$ & 45.0 & 30.0 & 45.0 & 20.0 & 52.5 & 65.0 & 30.0 & 36.2 \\
        Grok Imagine$^*$ & \textbf{67.5} & \textbf{55.0} & 50.0 & \textbf{45.0} & 47.5 & 75.0 & \textbf{55.0} & \textbf{57.5} \\
        \bottomrule
    \end{tabular}
\end{table*}

\begin{table*}[htbp]
    \centering
    \caption{\textbf{Temporal Sequencing — per-sub-category results (\%).} Full APEO breakdown for each model on every sub-category within this primitive. \textbf{Bold}: best per column within each sub-category.}
    \label{tab:app_detail_temporal_sequencing}
    \small
    \renewcommand{\arraystretch}{1.15}
    \begin{tabular}{@{} l cccccccc @{}}
        \toprule
        \textbf{Model} & $A_s$ & $A_p$ & $P_s$ & $P_p$ & $E_s$ & $E_p$ & $O_p$ & \textbf{APEO} \\
        \midrule
        \multicolumn{9}{l}{\textit{Overtake Hard (AD)}} \\
        \midrule
        CogVideoX1.5 & 38.3 & 3.3 & 91.7 & 3.3 & 45.0 & 96.7 & 3.3 & 26.7 \\
        Wan2.2 & 48.3 & 10.0 & 88.3 & 0.0 & 63.3 & 93.3 & 10.0 & 28.3 \\
        HunyuanVideo1.5 & 31.7 & 0.0 & \textbf{100.0} & 0.0 & 71.7 & \textbf{100.0} & 0.0 & 25.0 \\
        Cosmos-Predict2 & 18.3 & 3.3 & 83.3 & 3.3 & 68.3 & 96.7 & 3.3 & 26.7 \\
        \midrule
        Seedance 1.5$^*$ & 26.7 & 10.0 & 85.0 & 3.3 & 53.3 & 76.7 & 10.0 & 25.0 \\
        Kling 3.0$^*$ & 41.7 & 3.3 & 85.0 & 3.3 & \textbf{75.0} & 86.7 & 3.3 & 24.2 \\
        Seedance 2.0$^*$ & 45.0 & 6.7 & 91.7 & 3.3 & 73.3 & 60.0 & 3.3 & 18.3 \\
        Veo 3.1$^*$ & \textbf{73.3} & \textbf{50.0} & 88.3 & \textbf{30.0} & \textbf{75.0} & 90.0 & \textbf{46.7} & \textbf{54.2} \\
        Grok Imagine$^*$ & 58.3 & 26.7 & 85.0 & 20.0 & \textbf{75.0} & 93.3 & 26.7 & 41.7 \\
        \midrule
        \multicolumn{9}{l}{\textit{Overtake Accel (AD)}} \\
        \midrule
        CogVideoX1.5 & 36.7 & 20.0 & \textbf{80.0} & 20.0 & 50.0 & 93.3 & 26.7 & 40.0 \\
        Wan2.2 & 43.3 & 13.3 & 63.3 & 20.0 & 56.7 & 93.3 & 13.3 & 35.0 \\
        HunyuanVideo1.5 & 26.7 & 0.0 & 66.7 & 0.0 & 66.7 & 86.7 & 0.0 & 21.7 \\
        Cosmos-Predict2 & 10.0 & 0.0 & 43.3 & 0.0 & 43.3 & \textbf{100.0} & 0.0 & 25.0 \\
        \midrule
        Seedance 1.5$^*$ & 20.0 & 0.0 & 56.7 & 0.0 & 43.3 & 80.0 & 0.0 & 20.0 \\
        Kling 3.0$^*$ & 40.0 & 6.7 & 46.7 & 6.7 & 63.3 & 93.3 & 6.7 & 28.3 \\
        Seedance 2.0$^*$ & \textbf{60.0} & \textbf{33.3} & 70.0 & \textbf{33.3} & 73.3 & 86.7 & \textbf{40.0} & \textbf{48.3} \\
        Veo 3.1$^*$ & 46.7 & 0.0 & 46.7 & 0.0 & \textbf{83.3} & 80.0 & 0.0 & 20.0 \\
        Grok Imagine$^*$ & 40.0 & 0.0 & 50.0 & 0.0 & 60.0 & 86.7 & 0.0 & 21.7 \\
        \midrule
        \multicolumn{9}{l}{\textit{Overtake Brake (AD)}} \\
        \midrule
        CogVideoX1.5 & 30.0 & 20.0 & 70.0 & 0.0 & 46.7 & \textbf{100.0} & 20.0 & 35.0 \\
        Wan2.2 & 33.3 & 20.0 & 76.7 & 13.3 & 66.7 & 86.7 & 20.0 & 35.0 \\
        HunyuanVideo1.5 & 16.7 & 6.7 & \textbf{86.7} & 6.7 & 73.3 & 93.3 & 6.7 & 28.3 \\
        Cosmos-Predict2 & 16.7 & 6.7 & 43.3 & 6.7 & 83.3 & \textbf{100.0} & 6.7 & 30.0 \\
        \midrule
        Seedance 1.5$^*$ & 20.0 & 13.3 & 70.0 & 20.0 & 63.3 & 46.7 & 20.0 & 25.0 \\
        Kling 3.0$^*$ & 46.7 & 26.7 & 63.3 & 13.3 & 70.0 & 86.7 & 26.7 & 38.3 \\
        Seedance 2.0$^*$ & 46.7 & \textbf{33.3} & 83.3 & \textbf{33.3} & 73.3 & 66.7 & \textbf{40.0} & \textbf{43.3} \\
        Veo 3.1$^*$ & 50.0 & 6.7 & 56.7 & 0.0 & \textbf{90.0} & 93.3 & 6.7 & 26.7 \\
        Grok Imagine$^*$ & \textbf{53.3} & 20.0 & 53.3 & 13.3 & 86.7 & \textbf{100.0} & 20.0 & 38.3 \\
        \midrule
        \multicolumn{9}{l}{\textit{Gripper Timing (Rob)}} \\
        \midrule
        CogVideoX1.5 & 15.9 & 9.1 & 6.8 & 4.5 & 4.5 & 9.1 & 9.1 & 8.0 \\
        Wan2.2 & 56.8 & 18.2 & 47.7 & 18.2 & 15.9 & 13.6 & \textbf{36.4} & 21.6 \\
        HunyuanVideo1.5 & 52.3 & 27.3 & 40.9 & 4.5 & 29.5 & 31.8 & 4.5 & 17.0 \\
        Cosmos-Predict2 & 77.3 & 45.5 & 52.3 & 31.8 & 54.5 & 68.2 & 22.7 & 42.0 \\
        \midrule
        Seedance 1.5$^*$ & 54.5 & 9.1 & 43.2 & 9.1 & 29.5 & 13.6 & 13.6 & 11.4 \\
        Kling 3.0$^*$ & 77.3 & \textbf{59.1} & 75.0 & \textbf{36.4} & 68.2 & 59.1 & 31.8 & \textbf{46.6} \\
        Seedance 2.0$^*$ & 72.7 & 36.4 & 77.3 & 22.7 & 65.9 & 54.5 & \textbf{36.4} & 37.5 \\
        Veo 3.1$^*$ & 63.6 & 36.4 & \textbf{88.6} & 31.8 & \textbf{86.4} & \textbf{90.9} & 22.7 & 45.5 \\
        Grok Imagine$^*$ & \textbf{79.5} & 40.9 & 65.9 & \textbf{36.4} & 61.4 & 63.6 & 18.2 & 39.8 \\
        \bottomrule
    \end{tabular}
\end{table*}

\begin{table*}[htbp]
    \centering
    \caption{\textbf{Surface Friction — per-sub-category results (\%).} Full APEO breakdown for each model on every sub-category within this primitive. \textbf{Bold}: best per column within each sub-category.}
    \label{tab:app_detail_surface_friction}
    \small
    \renewcommand{\arraystretch}{1.15}
    \begin{tabular}{@{} l cccccccc @{}}
        \toprule
        \textbf{Model} & $A_s$ & $A_p$ & $P_s$ & $P_p$ & $E_s$ & $E_p$ & $O_p$ & \textbf{APEO} \\
        \midrule
        \multicolumn{9}{l}{\textit{Road Friction (AD)}} \\
        \midrule
        CogVideoX1.5 & 36.7 & 33.3 & 20.0 & 6.7 & 23.3 & 13.3 & 6.7 & 15.0 \\
        Wan2.2 & 23.3 & 26.7 & 13.3 & 6.7 & 50.0 & 6.7 & 6.7 & 11.7 \\
        HunyuanVideo1.5 & 76.7 & 73.3 & 40.0 & 13.3 & 26.7 & 13.3 & 20.0 & 30.0 \\
        Cosmos-Predict2 & 36.7 & 26.7 & 13.3 & 0.0 & 20.0 & 13.3 & 0.0 & 10.0 \\
        \midrule
        Seedance 1.5$^*$ & 76.7 & 80.0 & 46.7 & 20.0 & 40.0 & 20.0 & 20.0 & 35.0 \\
        Kling 3.0$^*$ & 83.3 & \textbf{93.3} & 50.0 & 20.0 & \textbf{96.7} & \textbf{100.0} & 20.0 & \textbf{58.3} \\
        Seedance 2.0$^*$ & 83.3 & 80.0 & 50.0 & 13.3 & 90.0 & 80.0 & 13.3 & 46.7 \\
        Veo 3.1$^*$ & 76.7 & 73.3 & 40.0 & 6.7 & 43.3 & 20.0 & 6.7 & 26.7 \\
        Grok Imagine$^*$ & \textbf{89.3} & 85.7 & \textbf{64.3} & \textbf{42.9} & 60.7 & 35.7 & \textbf{42.9} & 51.8 \\
        \midrule
        \multicolumn{9}{l}{\textit{Table Friction (Rob)}} \\
        \midrule
        CogVideoX1.5 & 6.7 & 0.0 & 6.7 & 0.0 & 6.7 & 40.0 & 0.0 & 10.0 \\
        Wan2.2 & 40.0 & 20.0 & 73.3 & 6.7 & 50.0 & 66.7 & 6.7 & 25.0 \\
        HunyuanVideo1.5 & 26.7 & 20.0 & 70.0 & 0.0 & 40.0 & 66.7 & 0.0 & 21.7 \\
        Cosmos-Predict2 & 46.7 & 26.7 & 20.0 & 0.0 & 20.0 & 66.7 & 0.0 & 23.3 \\
        \midrule
        Seedance 1.5$^*$ & 56.7 & 20.0 & \textbf{80.0} & 6.7 & \textbf{60.0} & 60.0 & 6.7 & 23.3 \\
        Kling 3.0$^*$ & 36.7 & 0.0 & 36.7 & 6.7 & 33.3 & 40.0 & 20.0 & 16.7 \\
        Seedance 2.0$^*$ & 53.3 & 13.3 & 70.0 & 13.3 & 53.3 & 46.7 & 26.7 & 25.0 \\
        Veo 3.1$^*$ & 56.7 & 33.3 & 50.0 & \textbf{26.7} & 53.3 & \textbf{86.7} & 20.0 & 41.7 \\
        Grok Imagine$^*$ & \textbf{60.0} & \textbf{40.0} & 56.7 & \textbf{26.7} & 56.7 & 80.0 & \textbf{40.0} & \textbf{46.7} \\
        \bottomrule
    \end{tabular}
\end{table*}

\begin{table*}[htbp]
    \centering
    \caption{\textbf{Material / Medium — per-sub-category results (\%).} Full APEO breakdown for each model on every sub-category within this primitive. \textbf{Bold}: best per column within each sub-category.}
    \label{tab:app_detail_materialmedium}
    \small
    \renewcommand{\arraystretch}{1.15}
    \begin{tabular}{@{} l cccccccc @{}}
        \toprule
        \textbf{Model} & $A_s$ & $A_p$ & $P_s$ & $P_p$ & $E_s$ & $E_p$ & $O_p$ & \textbf{APEO} \\
        \midrule
        \multicolumn{9}{l}{\textit{Road Medium (AD)}} \\
        \midrule
        CogVideoX1.5 & 53.3 & 53.3 & 50.0 & 0.0 & 80.0 & 80.0 & 0.0 & 33.3 \\
        Wan2.2 & 56.7 & 60.0 & 50.0 & 6.7 & 56.7 & 73.3 & 6.7 & 36.7 \\
        HunyuanVideo1.5 & 53.3 & 40.0 & 50.0 & 13.3 & 36.7 & 80.0 & 13.3 & 36.7 \\
        Cosmos-Predict2 & 53.3 & 40.0 & 40.0 & 0.0 & 63.3 & 46.7 & 0.0 & 21.7 \\
        \midrule
        Seedance 1.5$^*$ & 63.3 & 73.3 & 60.0 & 20.0 & 66.7 & 86.7 & 20.0 & 50.0 \\
        Kling 3.0$^*$ & 53.3 & 73.3 & 50.0 & 0.0 & 86.7 & 80.0 & 0.0 & 38.3 \\
        Seedance 2.0$^*$ & 53.3 & 73.3 & 53.3 & 6.7 & \textbf{90.0} & 86.7 & 6.7 & 43.3 \\
        Veo 3.1$^*$ & \textbf{86.7} & \textbf{80.0} & \textbf{83.3} & \textbf{73.3} & 73.3 & 53.3 & \textbf{73.3} & \textbf{70.0} \\
        Grok Imagine$^*$ & 71.4 & 57.1 & 67.9 & 35.7 & 85.7 & \textbf{92.9} & 42.9 & 57.1 \\
        \midrule
        \multicolumn{9}{l}{\textit{Deformability (Rob)}} \\
        \midrule
        CogVideoX1.5 & 10.0 & 0.0 & 0.0 & 0.0 & 0.0 & 26.7 & 0.0 & 6.7 \\
        Wan2.2 & 26.7 & 13.3 & 23.3 & 6.7 & 20.0 & 60.0 & 6.7 & 21.7 \\
        HunyuanVideo1.5 & 36.7 & 20.0 & 40.0 & 13.3 & 36.7 & 60.0 & 26.7 & 30.0 \\
        Cosmos-Predict2 & 66.7 & 33.3 & 43.3 & 26.7 & 56.7 & \textbf{80.0} & 26.7 & 41.7 \\
        \midrule
        Seedance 1.5$^*$ & 63.3 & 26.7 & 53.3 & 26.7 & 60.0 & \textbf{80.0} & 40.0 & 43.3 \\
        Kling 3.0$^*$ & 63.3 & 26.7 & 56.7 & 26.7 & 63.3 & 66.7 & 26.7 & 36.7 \\
        Seedance 2.0$^*$ & 50.0 & 26.7 & 53.3 & 26.7 & \textbf{70.0} & 60.0 & 26.7 & 35.0 \\
        Veo 3.1$^*$ & 60.0 & \textbf{60.0} & 70.0 & \textbf{46.7} & \textbf{70.0} & 66.7 & \textbf{60.0} & \textbf{58.3} \\
        Grok Imagine$^*$ & \textbf{80.0} & 40.0 & \textbf{73.3} & 40.0 & 66.7 & 53.3 & 53.3 & 46.7 \\
        \bottomrule
    \end{tabular}
\end{table*}

\begin{table*}[htbp]
    \centering
    \caption{\textbf{Obstacle Configuration — per-sub-category results (\%).} Full APEO breakdown for each model on every sub-category within this primitive. \textbf{Bold}: best per column within each sub-category.}
    \label{tab:app_detail_obstacle_configuration}
    \small
    \renewcommand{\arraystretch}{1.15}
    \begin{tabular}{@{} l cccccccc @{}}
        \toprule
        \textbf{Model} & $A_s$ & $A_p$ & $P_s$ & $P_p$ & $E_s$ & $E_p$ & $O_p$ & \textbf{APEO} \\
        \midrule
        \multicolumn{9}{l}{\textit{Scene Layout (AD)}} \\
        \midrule
        CogVideoX1.5 & 60.0 & 26.7 & 46.7 & 0.0 & 23.3 & 13.3 & 6.7 & 11.7 \\
        Wan2.2 & 53.3 & 20.0 & 43.3 & 0.0 & 53.3 & 13.3 & 6.7 & 10.0 \\
        HunyuanVideo1.5 & 63.3 & 26.7 & 50.0 & 0.0 & 23.3 & 13.3 & 0.0 & 10.0 \\
        Cosmos-Predict2 & 26.7 & 0.0 & 26.7 & 0.0 & \textbf{76.7} & \textbf{73.3} & 0.0 & 18.3 \\
        \midrule
        Seedance 1.5$^*$ & 83.3 & 66.7 & 60.0 & 20.0 & 63.3 & 40.0 & 33.3 & 40.0 \\
        Kling 3.0$^*$ & 86.7 & 73.3 & 60.0 & 26.7 & 73.3 & 60.0 & 26.7 & 46.7 \\
        Seedance 2.0$^*$ & 76.7 & 53.3 & 56.7 & 13.3 & 70.0 & 33.3 & 13.3 & 28.3 \\
        Veo 3.1$^*$ & 70.0 & 40.0 & 53.3 & 20.0 & 66.7 & 20.0 & 46.7 & 31.7 \\
        Grok Imagine$^*$ & \textbf{93.3} & \textbf{86.7} & \textbf{80.0} & \textbf{66.7} & 66.7 & 40.0 & \textbf{80.0} & \textbf{68.3} \\
        \midrule
        \multicolumn{9}{l}{\textit{Object Mass (Rob)}} \\
        \midrule
        CogVideoX1.5 & 10.0 & 6.7 & 3.3 & 0.0 & 3.3 & 33.3 & 0.0 & 10.0 \\
        Wan2.2 & 33.3 & 26.7 & 40.0 & 20.0 & 36.7 & 53.3 & 0.0 & 25.0 \\
        HunyuanVideo1.5 & 33.3 & 33.3 & 40.0 & 0.0 & 43.3 & 60.0 & 0.0 & 23.3 \\
        Cosmos-Predict2 & 40.0 & \textbf{46.7} & 36.7 & 13.3 & 40.0 & 46.7 & 6.7 & 28.3 \\
        \midrule
        Seedance 1.5$^*$ & 36.7 & 6.7 & 20.0 & 6.7 & 26.7 & 40.0 & 6.7 & 15.0 \\
        Kling 3.0$^*$ & 43.3 & 33.3 & 30.0 & 13.3 & 26.7 & 46.7 & 0.0 & 23.3 \\
        Seedance 2.0$^*$ & 53.3 & 20.0 & 46.7 & 0.0 & \textbf{70.0} & 53.3 & 0.0 & 18.3 \\
        Veo 3.1$^*$ & 66.7 & 33.3 & \textbf{60.0} & \textbf{33.3} & 63.3 & \textbf{66.7} & \textbf{26.7} & \textbf{40.0} \\
        Grok Imagine$^*$ & \textbf{80.0} & \textbf{46.7} & 46.7 & \textbf{33.3} & 56.7 & 40.0 & \textbf{26.7} & 36.7 \\
        \bottomrule
    \end{tabular}
\end{table*}

\begin{table*}[htbp]
    \centering
    \caption{\textbf{Representative VLM judge observations by failure mode.} Each row shows one verbatim excerpt illustrating the failure pattern. Observations are lightly truncated for space.}
    \label{tab:judge_examples}
    \small
    \renewcommand{\arraystretch}{1.3}
    \begin{tabular}{@{} p{1cm} p{1.3cm} p{10.5cm} @{}}
        \toprule
        \textbf{Dim.} & \textbf{Mode} & \textbf{Judge Observation (verbatim excerpt)} \\
        \midrule
        $A_s$ & Action Omission & \textit{The spatial error sub-type is Destination Accuracy. In Video A, the arm is supposed to pick up a donut and place it on the red plate, but the arm morphs massively, blocking the camera view, and fails to execute the task. In Video B, the arm is supposed to place the donut on the b} \\[3pt]
        $A_p$ & Baseline Inconsist. & \textit{There is no baseline consistency between the two videos. The arm kinematics and trajectories are completely different from the start, and neither video successfully demonstrates the intended destination accuracy task. The spatial difference is not clearly or unambiguously observa} \\[3pt]
        $P_s$ & Object Morphing & \textit{In Video A, the robotic arm exhibits severe morphing and kinematic violations, completely losing its structural integrity and blocking the camera view. In Video B, a donut spontaneously spawns mid-video, and a duplicate donut appears during the grasp, violating physical realism a} \\[3pt]
        $P_p$ & Mode Collapse & \textit{The pre-divergence trajectories are not identical. The arms move completely differently from the beginning, with Video A's arm morphing and blocking the view, while Video B's arm moves to grasp a spontaneously spawned object. There is no valid trajectory bifurcation alignment.} \\[3pt]
        $P_p$ & Causal Inversion & \textit{Although the initial gap distances are identical in the first frame, the trajectory bifurcation in Video B does not result in a decreasing gap as required; the gap increases instead.} \\[3pt]
        $E_s$ & Background Collapse & \textit{In Video A, the target object (donut) is missing from the start frame. The arm morphs massively, blocking the camera and destroying embodiment stability. In Video B, the donut is also missing from the start frame. A donut spontaneously spawns mid-video (unscripted object spawn), } \\[3pt]
        $E_p$ & Envir. Divergence & \textit{Due to the severe camera shift and the spontaneous appearance of the wall logo in Video B, the workspace background and camera angle are completely inconsistent between the two videos.} \\[3pt]
        $O_p$ & Mode Collapse & \textit{The expected outcome is that Video A successfully grasps and lifts the dark cap, while Video B misses. However, in both videos, the dark cap remains on the bed. Video A fails to grasp and lift the cap, resulting in mode collapse where both videos show the same final object state } \\[3pt]
        $O_p$ & Causal Inversion & \textit{At no point does the gap in Video B become smaller than in Video A. A causal inversion occurs where the gap in Video B is much larger than in Video A by the end of the clip, meaning the intended outcome failed completely.} \\[3pt]
        $O_p$ & Upstream Propag. & \textit{The expected outcome is that Video A shows the donut placed on the red plate, while Video B shows it placed on the bare table. Neither video shows the donut being placed at any destination, resulting in a complete failure to produce the expected state divergence.} \\[3pt]
        \bottomrule
    \end{tabular}
\end{table*}
\subsection{VLM Judge Example Observations}
\label{app:judge_examples}

Tables~\ref{tab:app_detail_spatial_alignment_ad}--\ref{tab:app_detail_obstacle_configuration} provide the full APEO breakdown for every model on every sub-category within each primitive.
These tables serve as the detailed backing data for the aggregated per-primitive tables in the main text (Tables~\ref{tab:prim_spatial_alignment}--\ref{tab:prim_obstacle_configuration}).

Each table is organized by sub-category (shown as italic headers), with models sorted by their overall APEO on the primitive.
Bold values indicate the best score per column within each sub-category.
Reading across a sub-category block reveals the full causal pipeline for each model on that specific task: which models can execute the Adherence ($A$), which produce correct differential physics ($P$), which maintain scene consistency ($E$), and which ultimately achieve the correct outcome ($O_p$).

To give the reader a concrete sense of what the VLM judge actually sees and reports, Table~\ref{tab:judge_examples} shows one representative observation for each failure mode identified in Section~\ref{sec:failure_analysis}.
All excerpts are verbatim from the judge's chain-of-thought reasoning, lightly truncated for space.

A few things stand out from these examples.
First, the judge's language is highly specific---it does not simply say ``the video is bad'' but identifies the exact mechanism of failure (``the arm morphs massively,'' ``the gap increases instead of decreasing,'' ``a donut spontaneously spawns mid-video'').
This specificity is what makes the APEO decomposition possible: the judge can distinguish between an action that was never executed ($A$ failure), physics that was individually plausible but causally meaningless ($P_p$ mode collapse), and an environment that hallucinated divergent artifacts ($E_p$ divergence).

Second, the failure descriptions are qualitatively different between single and paired modes.
Single-video failures describe generation-quality problems (morphing, teleportation, spawning), which are the kinds of artifacts any video quality metric could catch.
Paired failures describe \emph{relational} problems---the two videos are individually fine but fail to differ in the right way.
This distinction validates the core premise of our benchmark: paired evaluation tests a fundamentally different capability than single-video evaluation.

\subsection{Effect of Outcome Hints in the Generation Prompt}
\label{app:hint_ablation}

Our main protocol (Section~\ref{sec:construction}) deliberately withholds the 
physical outcome from the generation prompt, so that the model must simulate 
the consequence of the intervention from $x_0$ and $v$ rather than retrieve 
it from textual priors. To probe how much of the contrastive bottleneck 
arises from this design choice---specifically, to decompose paired-mode 
failure into outcome \emph{prediction} (can the model infer what should 
happen?) versus outcome \emph{rendering} (can it produce the prescribed 
transition once told?)---we run a controlled ablation on Grok-generated 
videos in which the generation prompt itself is augmented with an explicit 
outcome statement.

We compare two configurations: (1)~\textbf{Base}, the standard protocol 
where prompts contain only camera perspective, scene state, and intervened 
action (no outcome wording), and (2)~\textbf{+Hint}, where the same prompts 
are extended with a brief clause describing the intended physical outcome 
(e.g., \emph{``...causing the gap between the ego car and the leading 
vehicle to increase''}). The judge configuration and all evaluation procedures are held 
identical across the two conditions. The +Hint condition is evaluated across 
17 shared category splits, and we compare it against 
Grok's full-benchmark Base scores reported in Table~\ref{tab:main_results}.

Including the outcome in the generation prompt produces consistent but 
\emph{modest} gains across all four dimensions ($+0.047$ on average), with 
the largest improvements on Outcome Divergence ($+0.069$) and Physical 
Realism ($+0.050$). Two observations follow.

\paragraph{The hint does not close the gap.} Even when the model is 
explicitly told what should happen, paired Outcome Divergence rises only to 
$52.0\%$ and paired Physical Realism only to $47.6\%$. If the contrastive 
bottleneck were primarily a prediction failure---i.e., the model knew how 
to render a hard brake but could not infer that hard braking was the 
intended outcome---we would expect a much larger jump under +Hint. The 
modest magnitude of $\Delta$ indicates that a substantial portion of the gap 
reflects the model's inability to \emph{render} the prescribed physical 
transition correctly, regardless of whether that transition is stated 
explicitly in the prompt.

\definecolor{promptbg}{RGB}{245,245,250}
\definecolor{dimA}{RGB}{0,100,180}
\definecolor{dimP}{RGB}{180,80,0}
\definecolor{dimO}{RGB}{0,140,60}
\definecolor{dimE}{RGB}{140,0,140}

\newtcolorbox{promptbox}[1][]{
  colback=promptbg, colframe=black!60,
  fonttitle=\bfseries\small, title={#1},
  boxrule=0.4pt, arc=2pt, left=6pt, right=6pt, top=4pt, bottom=4pt,
  fontupper=\small
}

\section{VLM Judge Evaluation Prompt Template}
\label{app:eval_prompt}

We evaluate each counterfactual video pair using a Vision--Language Model (VLM) judge with a structured system prompt. The prompt follows a unified template instantiated per category. Below we present the generalized template, followed by the category-specific instantiation for one representative category (Category~1: Ego Following Speed Change).

\begin{table}[htbp]
\centering
\caption{Paired-comparison scores for \textbf{Grok} under \textbf{Base} 
(outcome withheld from generation prompt; full-benchmark scores from 
Table~\ref{tab:main_results}) vs.\ \textbf{+Hint} (outcome included in 
generation prompt). The VLM judge is identical in both conditions; only the 
generation prompt differs. $\Delta$ denotes absolute change.}
\label{tab:hint_ablation}
\begin{tabular}{lccc}
\toprule
\textbf{Dimension} & \textbf{Base} & \textbf{+Hint} & \textbf{$\Delta$} \\
\midrule
Adherence ($A_p$)                  & 0.492 & 0.535 & $+0.043$ \\
Physical Realism ($P_p$)           & 0.426 & 0.476 & $+0.050$ \\
Environmental Consistency ($E_p$)  & 0.697 & 0.725 & $+0.028$ \\
Outcome Divergence ($O_p$)         & 0.451 & 0.520 & $+0.069$ \\
\midrule
\textbf{Average}                   & 0.517 & 0.564 & $+0.047$ \\
\bottomrule
\end{tabular}
\end{table}
\subsection{Generalized Template Structure}
\label{app:template_structure}

Each evaluation prompt consists of five sections:

\begin{enumerate}[leftmargin=*, itemsep=2pt]
  \item \textbf{Role Assignment \& Task Framing.} The VLM is instructed to act as an \emph{Autonomous Driving Video or Robotic Arm Evaluator and Causal Inference Specialist}, performing paired comparative analysis of two dashcam videos generated from diverging text prompts that differ in a single causal variable.

  \item \textbf{Scenario Context.} A category-specific block specifying: (a) the \emph{target entity} undergoing the causal intervention, (b) \emph{context entities} that must remain invariant, (c) the \emph{intervention} applied in Video~A vs.\ Video~B, and (d) \emph{causal constraints} dictating which entities may and may not change.

  \item \textbf{A-P-O-E Rubric.} Four orthogonal evaluation dimensions, each with single-video and/or paired-video sub-metrics, strict binary (0/1) scoring, exhaustive failure-trigger lists, and chain-of-thought (CoT) reasoning examples.

  \item \textbf{Scoring Logic.} A strict pass/fail protocol: Score~1 requires flawless execution; \emph{any} single violation (ghosting, semantic bleed, missing action, background instability) triggers Score~0.

  \item \textbf{Output Schema.} A fixed JSON schema requiring CoT observations and binary scores for each sub-metric.
\end{enumerate}

\subsection{A-P-O-E Evaluation Dimensions}
\label{app:apoe_dims}

\begin{promptbox}[Dimension Definitions (Generalized)]

\textbf{\textcolor{dimA}{[A] Intervention Adherence (The Cause)}} --- Does each video correctly depict its assigned causal intervention without spatial misalignment, temporal omission, or target confusion?

\smallskip
\begin{itemize}[leftmargin=*, itemsep=1pt]
  \item \emph{Single-Video}: Score each video independently on whether the instructed action is visible and correctly applied to the target entity.
  \item \emph{Paired}: Is there a visually discernible difference between Video~A and Video~B that reflects the intended causal manipulation?
  \item \emph{Failure triggers}: Action Omission, Target Misalignment, Temporal Lag, Pseudo-Action Illusion, Lateral Redirection.
\end{itemize}

\medskip
\textbf{\textcolor{dimP}{[P] Physics}} --- Does the causal transition follow physically plausible rigid-body kinematics and non-holonomic vehicle constraints?

\smallskip
\begin{itemize}[leftmargin=*, itemsep=1pt]
  \item \emph{Single-Video}: Check each video for kinematic violations, topology/ghosting artifacts, non-holonomic failures, morphing, directional inversion, and oscillatory motion.
  \item \emph{Paired (Trajectory Bifurcation Alignment)}: The initial segment before the intervention must show identical states; divergence should begin only at the intervention onset.
\end{itemize}

\medskip
\textbf{\textcolor{dimO}{[O] Counterfactual State Divergence (The Effect)}} --- Does the causally expected outcome manifest at any point during the video?

\smallskip
\begin{itemize}[leftmargin=*, itemsep=1pt]
  \item \emph{Paired only}: Assess whether the expected state difference (e.g., gap closure, lane position change, object appearance/disappearance) emerges at any point---not restricted to the final frame.
  \item \emph{Failure triggers}: Mode Collapse (superficial action with no effect), Causal Inversion (opposite outcome), No Delta (identical states throughout).
\end{itemize}

\medskip
\textbf{\textcolor{dimE}{[E] Contextual \& Embodiment Consistency (Environment / Ceteris Paribus)}} --- Do non-causal background variables maintain feature constancy?

\smallskip
\begin{itemize}[leftmargin=*, itemsep=1pt]
  \item \emph{Single-Video (Absolute Canvas Stability)}: Lane markings, buildings, road surface, bystander vehicle identities, and camera angle must remain stable throughout each video.
  \item \emph{Paired (Counterfactual Canvas Consistency)}: Masking out the target entity's changed behavior, the baseline road environment and bystander behaviors must be consistent across Video~A and Video~B.
  \item \emph{Failure triggers}: Background Collapse, Object Permanence Failure, Camera Hallucination, Phantom Road Geometry, Off-Road Excursion.
\end{itemize}

\end{promptbox}

\subsection{Output JSON Schema}
\label{app:json_schema}

The VLM judge is required to output a single JSON object with the following structure. Each block contains a free-text \texttt{observation} field (CoT reasoning) and binary score(s).

\begin{promptbox}[Required Output Schema]
\begin{verbatim}
{
  "A_Adherence_Single":  { "observation": "<CoT>",
                           "score_video_A": 0|1,
                           "score_video_B": 0|1 },
  "A_Adherence_Paired":  { "observation": "<CoT>",
                           "score": 0|1 },
  "P_Physics_Single":    { "observation": "<CoT>",
                           "score_video_A": 0|1,
                           "score_video_B": 0|1 },
  "P_Physics_Paired":    { "observation": "<CoT>",
                           "score": 0|1 },
  "O_Outcome_Paired":    { "observation": "<CoT>",
                           "score": 0|1 },
  "E_Environment_Single":{ "observation": "<CoT>",
                           "score_video_A": 0|1,
                           "score_video_B": 0|1 },
  "E_Environment_Paired":{ "observation": "<CoT>",
                           "score": 0|1 }
}
\end{verbatim}
\end{promptbox}

\subsection{Category-Specific Instantiation}
\label{app:cat_instantiation}

For each of theevaluation categories, the following placeholders in the template are filled with category-specific content:

\begin{promptbox}[Template Placeholders]
\textbf{Scenario Context:}
\begin{itemize}[leftmargin=*, nosep]
  \item \texttt{[TARGET\_ENTITY]} --- The entity undergoing the causal intervention (e.g., ego car, traffic light, pedestrian).
  \item \texttt{[CONTEXT\_ENTITIES]} --- Entities that must remain invariant across both videos (e.g., front car, bystander vehicles, road infrastructure).
  \item \texttt{[INTERVENTION\_A]} --- Description of the baseline/control behavior shown in Video~A.
  \item \texttt{[INTERVENTION\_B]} --- Description of the causal manipulation shown in Video~B.
  \item \texttt{[CAUSAL\_CONSTRAINT]} --- Which entities may change and which must not.
  \item \texttt{[SCENARIO\_VARIANTS]} --- Notes on visual variants across samples within this category (e.g., different vehicle types) that do not affect evaluation logic.
\end{itemize}

\medskip
\textbf{Category-Specific Failure Triggers (per dimension):}
\begin{itemize}[leftmargin=*, nosep]
  \item \texttt{[A\_FAILURE\_TRIGGERS]} --- Adherence failures specific to this category's intervention type.
  \item \texttt{[P\_FAILURE\_TRIGGERS]} --- Physics/dynamics failures relevant to the physical motion involved.
  \item \texttt{[O\_FAILURE\_TRIGGERS]} --- Outcome failures describing what ``no causal effect'' looks like for this category.
  \item \texttt{[E\_FAILURE\_TRIGGERS]} --- Environment failures specific to the scene elements present.
\end{itemize}

\medskip
\textbf{CoT Examples:}
\begin{itemize}[leftmargin=*, nosep]
  \item \texttt{[COT\_EXAMPLES]} --- 2--4 chain-of-thought reasoning examples showing Pass and Fail cases for the most common ambiguities in this category.
\end{itemize}
\end{promptbox}

\subsection{Instantiation Example: Category~1 (Ego Following Speed Change)}
\label{app:cat1_example}

Below we show the complete instantiation for Category~1 as a concrete example.

\begin{promptbox}[Scenario Context --- Category 1]
\textbf{[TARGET\_ENTITY]:} The ego car (camera vehicle).

\smallskip
\textbf{[CONTEXT\_ENTITIES]:} A front car driving ahead in the same lane (may be a taxi, bus, sedan, van, or truck). One or more bystander vehicles in adjacent lanes or parked on the roadside. Bystander vehicles \emph{must remain unchanged} across both videos.

\smallskip
\textbf{[INTERVENTION\_A]:} The ego car maintains its current speed, keeping a constant following distance from the front car throughout the video.

\smallskip
\textbf{[INTERVENTION\_B]:} The ego car suddenly accelerates hard while remaining in its lane, closing the gap to the front car.

\smallskip
\textbf{[CAUSAL\_CONSTRAINT]:} The front car and all bystander vehicles must continue at their original speeds in both videos. Only the ego car's speed behavior changes.

\smallskip
\textbf{[SCENARIO\_VARIANTS]:} The front car type varies across samples (yellow taxi, orange bus, white bus, sedan, van). This does not affect evaluation logic.
\end{promptbox}

\begin{promptbox}[\texttt{[A\_FAILURE\_TRIGGERS]} --- Category 1]
\begin{itemize}[leftmargin=*, nosep]
  \item \emph{Action Omission}: In Video~A, the gap shows clearly obvious and substantial deviation from constant; in Video~B, no visible acceleration and gap does not decrease.
  \item \emph{Target Misalignment}: The front car speeds up instead of the ego car, creating an illusion of gap closure.
  \item \emph{Temporal Lag}: In Video~B, the ego car waits until the final frames to begin accelerating.
  \item \emph{Pseudo-Acceleration Illusion}: Apparent gap closure caused entirely by road curvature or camera zoom, not actual speed change.
  \item \emph{Lateral Redirection}: The ego car changes lane to gain speed rather than accelerating longitudinally.
  \item \emph{Scene-Flow Disambiguation}: When gap closure could be explained by front-car deceleration, use independent scene-flow cues (rate of road markings, ground texture, roadside objects passing the camera) as ego-car velocity proxy. No increase in scene-flow rate means the ego car did not accelerate.
\end{itemize}
\end{promptbox}

\begin{promptbox}[\texttt{[P\_FAILURE\_TRIGGERS]} --- Category 1]
\begin{itemize}[leftmargin=*, nosep]
  \item \emph{Kinematic Violation}: Instantaneous speed jump from current speed to high speed in a single frame (no ramp-up).
  \item \emph{Topology/Ghosting}: Ego car's front bumper phases through the rear of the front car without collision response.
  \item \emph{Non-Holonomic Failure}: Ego car slides laterally (drifts sideways) during acceleration.
  \item \emph{Morphing}: Ego car or front car visually distorts, stretches, or changes size during acceleration.
  \item \emph{Directional Inversion}: Any vehicle travels opposite to its established heading.
  \item \emph{Oscillatory Motion}: Repetitive forward--reverse--forward motion (generation artifact).
\end{itemize}
\end{promptbox}

\begin{promptbox}[\texttt{[O\_FAILURE\_TRIGGERS]} --- Category 1]
\begin{itemize}[leftmargin=*, nosep]
  \item \emph{Mode Collapse}: Ego car in Video~B shows acceleration cues (motion blur, engine sound) but the gap never meaningfully closes at any point.
  \item \emph{Causal Inversion}: Gap in Video~B is larger than in Video~A at all observable points.
  \item \emph{No Delta}: Front car appears the same angular size throughout both videos.
\end{itemize}
\end{promptbox}

\begin{promptbox}[\texttt{[E\_FAILURE\_TRIGGERS]} --- Category 1]
\begin{itemize}[leftmargin=*, nosep]
  \item \emph{Background Collapse}: Lane markings change count/style; buildings liquefy; road surface changes texture.
  \item \emph{Object Permanence Failure}: Front car transforms identity (taxi becomes bus); bystander vehicles change lane/speed.
  \item \emph{Camera Hallucination}: Dashcam angle or height changes abruptly mid-video.
  \item \emph{Phantom Road Geometry}: New lanes, intersections, or infrastructure materialise mid-video.
  \item \emph{Off-Road Excursion}: Any vehicle leaves paved road surface.
\end{itemize}
\end{promptbox}

\begin{promptbox}[\texttt{[COT\_EXAMPLES]} --- Category 1 (selected)]
\textbf{[A] Fail Example:}\\
\texttt{"observation": "In Video B, the front car (orange bus) appeared to grow larger, but this was because the bus itself slowed down --- the ego car's scene-flow cues did not increase. The acceleration was applied to the wrong entity."} $\rightarrow$ \texttt{score\_video\_B: 0}

\medskip
\textbf{[P] Fail Example:}\\
\texttt{"observation": "In Video B, the ego car executed a lane change into an adjacent lane to accelerate, rather than accelerating within its own lane. Changing lanes violates the single-lane following constraint."} $\rightarrow$ \texttt{score\_video\_B: 0}

\medskip
\textbf{[O] Fail Example:}\\
\texttt{"observation": "Throughout Video B, the front car maintained the same angular size as in Video A. At no point did the spatial gap noticeably close despite the ego car appearing to accelerate. The speed change produced no closing effect."} $\rightarrow$ \texttt{score: 0}

\medskip
\textbf{[E] Fail Example:}\\
\texttt{"observation": "New road geometry spontaneously appeared mid-video --- additional lanes materialised where there was previously grass. Phantom environment generation undermines the shared starting-state assumption."} $\rightarrow$ \texttt{score\_video\_B: 0}
\end{promptbox}

\end{document}